\documentclass[10pt,twocolumn,letterpaper]{article}

\usepackage[accsupp]{axessibility}  
\usepackage{iccv}
\usepackage{times}
\usepackage{epsfig}
\usepackage{graphicx}
\usepackage{amsmath}
\usepackage{amssymb}

\iccvfinalcopy 

\def\iccvPaperID{1632} 

\usepackage{graphicx}
\usepackage{amsmath}
\usepackage{amssymb}
\usepackage{booktabs}

\usepackage{caption}
\usepackage{subcaption}
\usepackage{outlines}
\usepackage[dvipsnames]{xcolor}
\usepackage{cuted}
\usepackage{bm}
\usepackage{array}
\usepackage{algpseudocode}
\usepackage{algorithm}
\usepackage{multirow}

\usepackage[pagebackref=true,breaklinks=true,letterpaper=true,colorlinks,bookmarks=false]{hyperref}

\usepackage[capitalize]{cleveref}
\crefname{section}{Sec.}{Secs.}
\Crefname{section}{Section}{Sections}
\Crefname{table}{Table}{Tables}
\crefname{table}{Tab.}{Tabs.}
\definecolor{Cerulean}{rgb}{0.0, 0.48, 0.65}
\newcommand{\rowNumber}[1]{\textcolor{Cerulean}{#1}}

\newcommand{\x}{\bm{\mathrm{x}}}
\newcommand{\zm}{\bm z^\textsf{m}}
\newcommand{\cm}{\bm c^\textsf{m}}
\newcommand{\Nstd}{\mathcal{N}(\bm 0, \bm I)}
\newcommand{\yes}{\checkmark}

\begin{document}

\title{\vspace{-0.3cm}StyleInV: A Temporal Style Modulated Inversion Network for \\ Unconditional Video Generation\vspace{-0.2cm}}

\author{Yuhan Wang, Liming Jiang, Chen Change Loy\\
S-Lab, Nanyang Technological University\\
{\tt\small \{yuhan004, liming002, ccloy\}@ntu.edu.sg}
}
\maketitle

\begin{abstract}
Unconditional video generation is a challenging task that involves synthesizing high-quality videos that are both coherent and of extended duration. To address this challenge, researchers have used pretrained StyleGAN image generators for high-quality frame synthesis and focused on motion generator design. The motion generator is trained in an autoregressive manner using heavy 3D convolutional discriminators to ensure motion coherence during video generation. In this paper, we introduce a novel motion generator design that uses a learning-based inversion network for GAN. The encoder in our method captures rich and smooth priors from encoding images to latents, and given the latent of an initially generated frame as guidance, our method can generate smooth future latent by modulating the inversion encoder temporally. Our method enjoys the advantage of sparse training and naturally constrains the generation space of our motion generator with the inversion network guided by the initial frame, eliminating the need for heavy discriminators. Moreover, our method supports style transfer with simple fine-tuning when the encoder is paired with a pretrained StyleGAN generator. Extensive experiments conducted on various benchmarks demonstrate the superiority of our method in generating long and high-resolution videos with decent single-frame quality and temporal consistency. Project website: \href{https://www.mmlab-ntu.com/project/styleinv/index.html}{https://www.mmlab-ntu.com/project/styleinv/index.html}.
\end{abstract}


\vspace{-0.1cm}
\section{Introduction}
\label{sec:intro}

\begin{figure}
    \centering
    \includegraphics[width=0.88\linewidth]{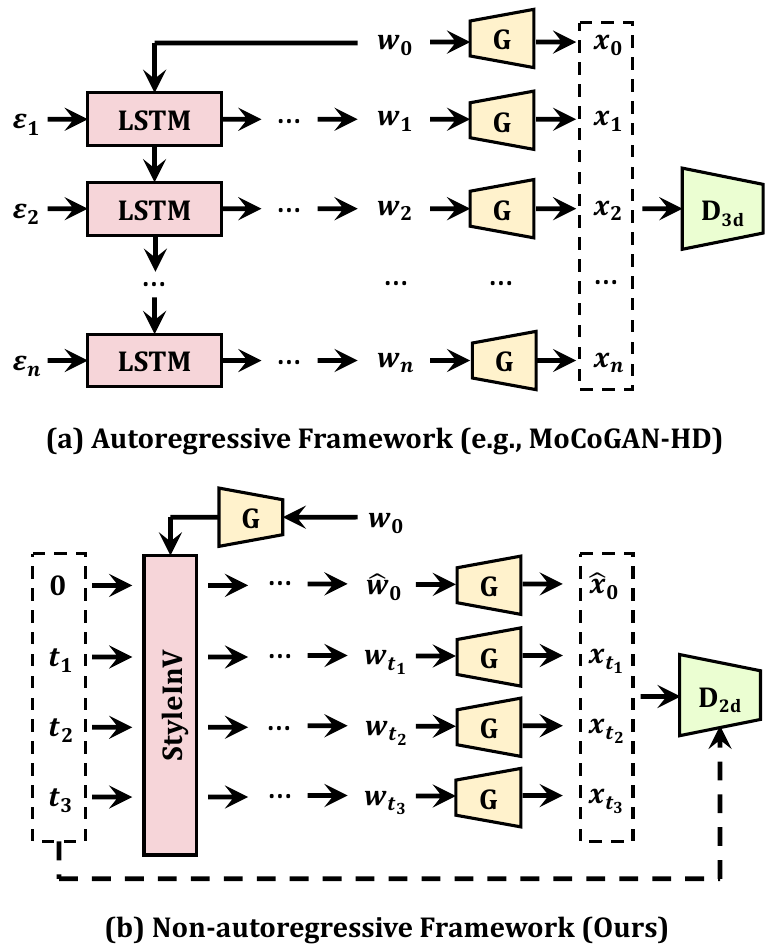}
    \caption{A comparison between autoregressive and non-autoregressive pipeline: (a) Previous autoregressive motion generators require generating the whole clip for a 3D-convolution-based discriminator. (b)  Our non-autoregressive motion generator, \textbf{StyleInV}, is an inversion network modulated by temporal style (as a random function of $t$), which enjoys sparse training using a 2D-convolution-based discriminator.}
    \label{fig:autoregressive}
\end{figure}
\begin{figure}
    \centering
    \includegraphics[width=0.9\linewidth]{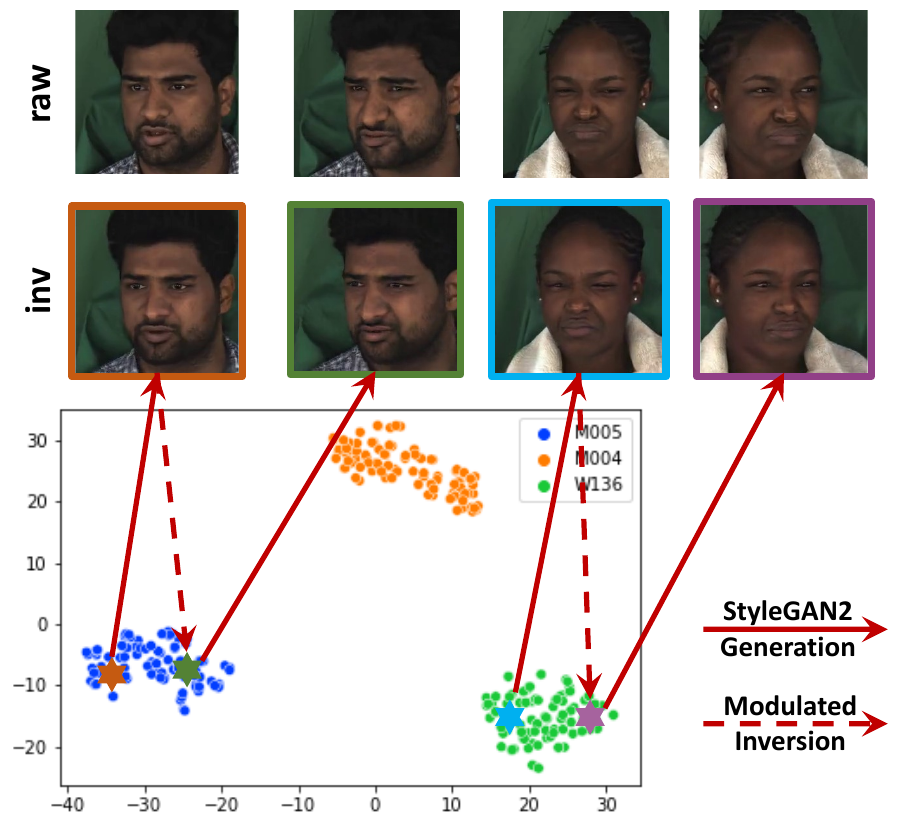}
    \caption{Inverted latent space visualization and modulated inversion process: When the StyleGAN generator is trained with video frame data, $\mathcal{W}$ space is well clustered by human identities and provides promising inversion results. Thus, the modulated inversion process can easily find the target latent corresponding to the same identity (shifted to the next motion) as the source one.}
    \label{fig:mod_inv}
\end{figure}

Unconditional video generation aims at learning a generative model to create novel videos from latent vectors.
Despite extensive studies~\cite{vgan, tgan, tganv2, mocogan, dvdgan, videogpt} in addressing this problem, it remains challenging to generate high-resolution videos with both favorable quality and motion coherence over a long-term duration. The core difficulties in this task lie in modeling consistent motion and managing the high memory consumption introduced by the addition of the temporal dimension.

To ensure high single-frame resolution and quality, many existing studies, such as MoCoGAN-HD~\cite{mocoganhd}, employ a powerful image generator such as StyleGAN~\cite{karras2019style} as a backbone to serve as a strong generative prior. This approach shifts the focus towards developing a robust motion generator that can capture temporally coherent motion. Most of these methods model motion in an auto-regressive manner, where the next latent is sampled conditioned on the previous one (see Fig.~\ref{fig:autoregressive}). However, this design has two main drawbacks. First, while good performance requires seeing a long sequence of images, the use of heavy 3D discriminators limits its ability to be trained with longer videos. Second, the autoregressive motion generator can lead to motion collapse when extrapolating to generate longer videos.

In this study, we present an effective framework for non-autoregressive motion generation that is capable of generating long and high-resolution videos. Our approach leverages learning-based Generative Adversarial Network (GAN) inversion, which learns the inverse mapping of GANs via an inversion network that consists of an encoder and a decoder\footnote{In many contexts, the decoder is a StyleGAN, and the encoder learns to encode a given image to meaningful latent vectors in the StyleGAN space. There is a variety of image manipulation applications~\cite{inversionsurvey,pSp, hyperstyle} developed based upon such an inversion framework.}. 
To generate long and coherent videos, we exploit the unique characteristic of the inversion encoder, which captures a rich and smooth manifold between the mapping of images and latent. As illustrated in Fig.~\ref{fig:autoregressive}, to generate a sequence of smooth motion latents, we just need to provide the initial latent code and modulate the inversion encoder with temporal style codes, which are encodings of timestamps with randomness. The motion latents can then be mapped by a StyleGAN decoder to generate a video.

The proposed framework offers several advantages in a single unified framework. \textbf{First}, the use of an inversion network naturally constrains the generation space to stay consistent with the desired appearance, which is defined by the initial latent code. As demonstrated in Fig.~\ref{fig:mod_inv}, this leads to a significant benefit.
\textbf{Second}, thanks to the flexibility of the inversion network in accepting temporal styles of arbitrary timestamps, the framework allows non-autoregressive generation and sparse training~\cite{digan, styleganv}. These merits help alleviate the need for heavy discriminators to ensure temporal consistency, as is required in existing approaches. In our implementation, we only need to use a 2D convolutional discriminator instead of a 3D discriminator like MoCoGAN-HD.
\textbf{Third}, Unlike existing state-of-the-art methods \cite{digan, styleganv, brooks2022longvideogan} that couple content and motion decoding in one synthesis network, our framework can naturally support content decoder fine-tuning on different image datasets. Specifically, after fine-tuning the decoder (\eg, StyleGAN2) on another image dataset with the mapping layers and low-resolution synthesis layers fixed, given the same sequence of synthesized motion latents, the generated video can possess the new style of the fine-tuning dataset while preserving the motion patterns of the video generated by the parent content decoder.

The main contribution of this work is a novel motion generator that modulates a GAN inversion network. This is the first attempt to build such a generator, and it offers several advantages in a unified framework over existing approaches. These advantages include consistent generation, sparse training, and flexibility in supporting style transfer with simple fine-tuning. 
We additionally contribute a reformulation to the conventional sparse training, through first-frame-aware acyclic positional encoding (FFA-APE) and first-frame-aware sparse training (FFA-ST), to ensure that our motion generator can faithfully reconstruct the initial frame and that the generated video is smooth and continuous.
Extensive experiments on DeeperForensics~\cite{jiang2020deeperforensics}, FaceForensics~\cite{faceforensics}, SkyTimelapse~\cite{skytimelapse} and Tai-Chi-HD~\cite{taichi} datasets show that our model is comparable to or even better than state-of-the-art unconditional video generation methods~\cite{mocoganhd, digan, styleganv} both qualitatively and quantitatively.
\section{Related Work}
\label{sec:related}

\noindent\textbf{GAN inversion.}
The goal of GAN inversion is to find the corresponding vector in the latent space of a pretrained GAN~\cite{karras2019style, karras2020analyzing} to reconstruct the input image. Existing methods can be classified into three categories~\cite{inversionsurvey}: (1) learning-based methods~\cite{L-inv-chai2021using, L-inv-e4e, L-inv-ghfeat, L-inv-restyle, L-inv-sam, L-inv-wei2021simple, pSp}, which leverage an encoder network to directly map an image into a latent vector; (2) optimization-based methods~\cite{O-inv-hijackgan, O-inv-styleflow, O-inv-stylespace, O-inv-xu2021continuity, O-inv-zhu2020improved, O-inv-zhuang2021enjoy}, which iteratively find the latent vector that best reconstructs the input image using gradient descent; and (3) hybrid models~\cite{H-inv-bau2019seeing, H-inv-bau2019semantic, H-inv-chai2021ensembling, H-inv-idinvert}, which initialize the iteration process with the result of an encoder network.
The design of our motion generator follows the learning-based approach. Therefore, our method is trainable, efficient for single-image inference, and suitable for hierarchical modulation. 
We devise the motion generator on the $\mathcal{W}$ space and use the StyleGAN generated latent as the initial content code to guide the modulated inversion process (see Fig.~\ref{fig:mod_inv}).

\noindent\textbf{Unconditional video generation.}
Unconditional video generation aims to model the distribution of real videos in a training dataset and generate videos from sampled noise vectors. Many recent studies on this topic are inspired by the success of GANs in image generation. VGAN~\cite{vgan} applies 3D convolutions in both the generator and discriminator, while TGAN~\cite{tgan} optimizes this design by decomposing the generator into an \textit{image generator}, which is shared by the generation of each frame, and a \textit{motion generator}\footnote{In the original paper of TGAN~\cite{tgan}, the authors called this module \emph{temporal generator}, which is equivalent to the \emph{motion generator} used in subsequent studies~\cite{mocogan, mocoganhd, digan} and in our paper.}. This framework has been followed by most subsequent studies, such as MoCoGAN~\cite{mocogan}, which applies a content-motion decomposition. Some approaches~\cite{tganv2, dvdgan, ldvdgan} have focused on reducing the computational cost of the video discriminator, but the cost is still proportional to the video duration and resolution.
Some recent methods have applied more advanced generative frameworks and techniques to unconditional video generation. For example, VideoGPT~\cite{videogpt} uses VQ-VAE~\cite{vq-vae-2} and GPT~\cite{image-gpt} to formulate a non-GAN-based video generation approach. Recent studies have also explored unconditional video generation with higher resolution and longer duration. For example, Long-Video-GAN~\cite{brooks2022longvideogan} develops a two-phase model that focuses on improving the long-term temporal dynamics of video generation. MoCoGAN-HD~\cite{mocoganhd} and StyleVideoGAN~\cite{stylevideogan} study the generation of latent trajectories in the latent space of a pretrained StyleGAN2 generator. Our approach is inspired by these studies, but differs in the design of the motion generator. Our motion generator is non-autoregressive, thus alleviating the use of heavy discriminators, and it is unique since it obtains the motion latent via modulating a GAN inversion network. This design allows us to attain better motion consistency and semantics.

Recent works ~\cite{digan, styleganv} explored neural representation-based generators and trained them sparsely as an image GAN. StyleSV \cite{zhang2022towards} improves this framework by introducing StyleGAN3~\cite{stylegan3} architecture and several temporal designs. In our work, we extend the idea of sparse training to first-frame-aware sparse training, allowing it to be applied to a generation pipeline conditioned on the initial latent.

\noindent\textbf{Diffusion-based video generation.} 
The diffusion models \cite{ddpm, stable-diffusion}, a new paradigm for image generation tasks, have also achieved significant progress in the task of unconditional video generation~\cite{video-diffusion,videofusion,MCVD,LGC-VD}. Despite their success, temporal consistency is still an open problem for diffusion models, and GAN-based models exhibit a clear advantage in terms of inference speed.

\section{Methodology}
\label{sec:method}

\begin{figure}
    \centering
    \includegraphics[width=0.99\linewidth]{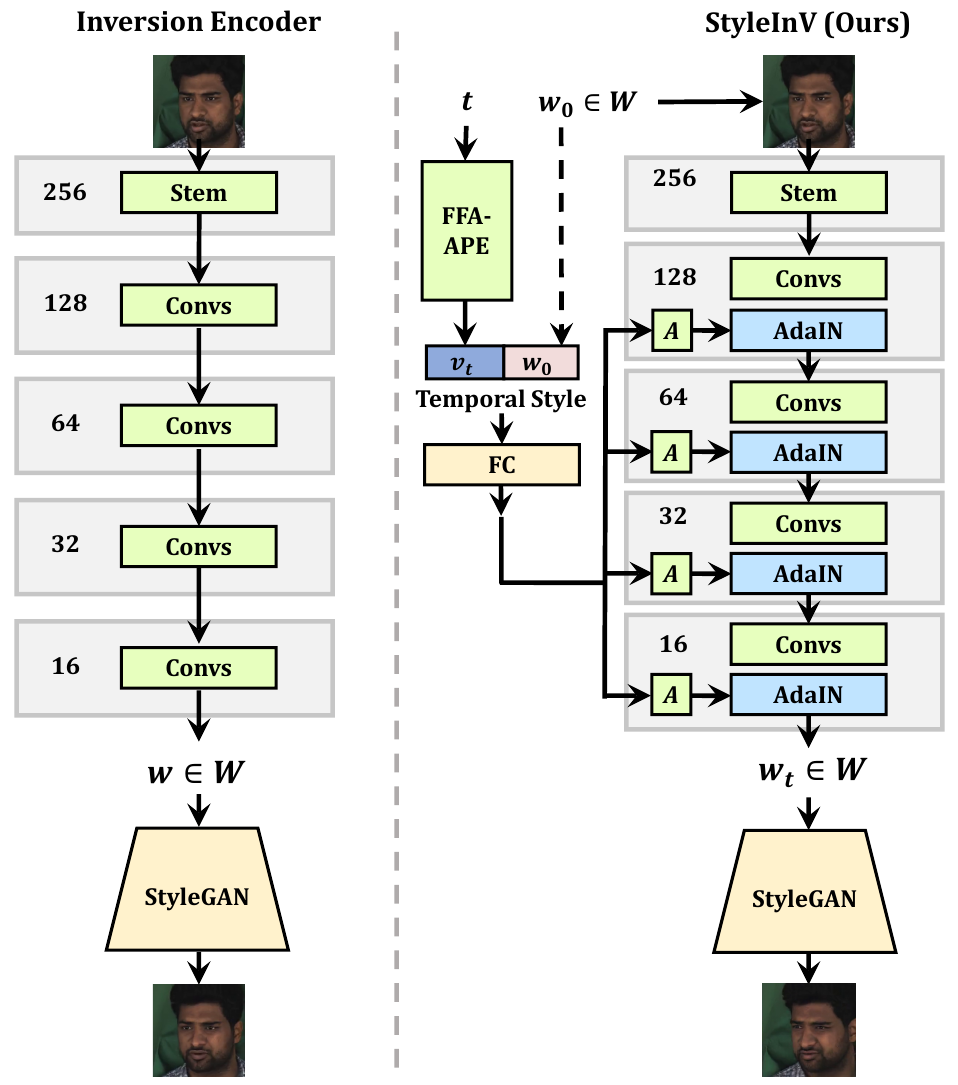}
    \caption{From a typical inversion encoder to StyleInV: We add \texttt{AdaIN} layers at the end of each residual block to inject the temporal style, which is a combination of time positional encoding and the first frame latent code. Here ``A'' stands for a learned affine transform \cite{karras2019style}.}
    \label{fig:pipeline}
\end{figure}

\subsection{Preliminaries of Inversion Encoder}

An inversion encoder maps an input image to a vector in the $\mathcal{W}$ or $\mathcal{W+}$ latent space of a pretrained StyleGAN2 generator. The generated image that corresponds to this vector should faithfully reconstruct the details of the input image. Therefore, when based on $\mathcal{W}$ latent space, given an input image $\bm{\mathrm{x}}$, the reconstruction process can be defined on top of the inversion network $\mathrm{Inv}$ as:
\begin{equation}
    \label{eq:inv}
    \hat{\x}:= G(\mathrm{Inv}(\x)) := G(E(\x) + \overline{\bm{\mathrm{w}}}).
\end{equation}
Here $E$ and $G$ denote the inversion encoder and StyleGAN generator, respectively. $\overline{\bm{\mathrm{w}}} \in \mathbb{R}^{512}$ denotes the average latent vector of the generator in the $\mathcal{W}$ latent space.
In our implementation, the encoder $E$ is a convolutional network backbone that outputs a 512-dimensional vector from the last layer embedding, as shown in Fig.~\ref{fig:pipeline}(left). We build the encoder on the $\mathcal{W}$ latent space, which eases the design of temporal modulation.

\subsection{Temporal Style Modulated Inversion Encoder}

We observe that the latent space of a StyleGAN trained on a video dataset is typically well-clustered by its content subject. Figure~\ref{fig:mod_inv} shows an example of human face videos, where we depict the results of inverting video clips of different identities into the $\mathcal{W}$ space and visualizing them with t-SNE~\cite{tsne}. 
It can be observed that the latent space is grouped by human identities. 
We also observe the same property in video datasets that follow other distributions. 
This phenomenon suggests that the inversion network inherits some important temporal priors that we could leverage to maintain motion consistency in generated videos. 

Motivated by this observation, we propose \textbf{StyleInV}, in which the motion latent is generated by modulating a GAN inversion network with temporal styles. 
Figure~\ref{fig:pipeline}(right) illustrates the pipeline of our framework. The temporal style $\bm{s}_t$ of a timestamp $t$ consists of two parts: the motion code $\bm{v}_t$ and the latent code of the initial frame $\bm{w}_0$. Inspired by~\cite{styleganv}, we use an acyclic positional encoding module to compute a dynamic embedding of the timestamp $t$. However, unlike~\cite{styleganv}, we make the embedding of the zero timestamp fixed, so this module becomes first-frame-aware. We provide more details in Section~\ref{sec:ffa-ape}. The latent code $\bm{w}_0$ of the initial frame is concatenated with the motion code for content-adaptive affine transform.

The temporal style is injected into the inversion encoder through \texttt{AdaIN} layers at the end of each convolution block. With this design, the encoder $E$ of StyleInV becomes a function of the initial latent code $w_0$ and timestamp $t$. The modulated inversion process can be defined as:
\begin{equation}
    \small
    \label{eq:styleinv}
    \hat{\x_t} := G(\mathrm{StyleInV}(\bm{w}_0,t)) := G(E(G(\bm{w}_0), \bm{s}_t) + \bm{w}_0).
\end{equation}
Notably, the output of $E$ serves as the residual w.r.t. $w_0$, instead of $\overline{\bm{\mathrm{w}}}$. This modification provides more explicit content information guidance for the inversion encoder. 

During training, we first train a raw inversion encoder following Eq.~\eqref{eq:inv} on all video frames. Then, we use this network to initialize the weights of all convolution layers in the StyleInV encoder. Other parameters (\eg, FFA-APE and Affine Transforms) are randomly initialized. Finally, the entire StyleInV encoder is trained end-to-end.

\subsection{FFA-APE}
\label{sec:ffa-ape}

\begin{figure}
    \centering
    \includegraphics[width=1.0\linewidth]{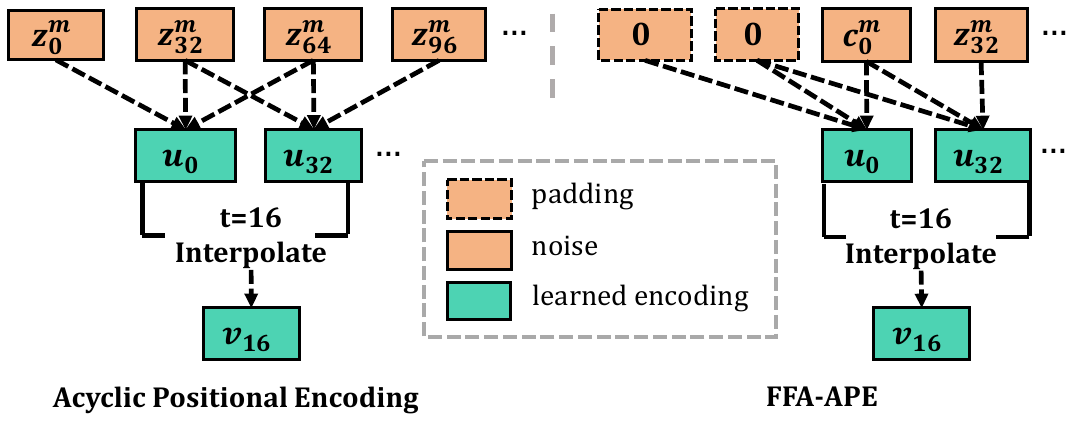}
    \caption{FFA-APE: A simplified case to compute $v_{16}$ for demonstrating the original acyclic positional encoding and our FFA-APE when $\delta^z=32$ and \texttt{conv1d} kernel size is $3$. In FFA-APE, the encoding of zero timestamp ($u_0$) only depends on constant paddings and a constant noise vector, thus is fixed for any sampled noise vector sequence.}
    \label{fig:ffa-ape}
\end{figure}

The original implementation of acyclic positional encoding (APE) \cite{styleganv} samples a series of noise vectors $\zm_{t_0},\cdots,\zm_{t_n} \sim \Nstd$ where $t_i=i\cdot \delta^z$. We call these temporal points \textit{anchor points}. Here $\delta^z$ is a set constant distance between adjacent anchor points. Then, the noise vectors are mapped to tokens $\bm u_{t_0},\cdots,\bm u_{t_n}$ by a padding-less \texttt{conv1d}-based motion mapping network. The computation of the acyclic positional encoding $\bm v_t$ of arbitrary timestamp $t$ is achieved by a scalable and learnable interpolation between the tokens of two adjacent \textit{anchor points} that cover $t$. The computation pipeline is shown in Fig.~\ref{fig:ffa-ape}. 

In our non-autoregressive generation pipeline, the modulated inversion encoder needs to faithfully reconstruct the initial frame when the input timestamp is zero, making it necessary to fix the computation of the APE for the zero timestamp $\bm v_0$. The original APE computation for $\bm v_0$ is dynamic and depends on randomly sampled noise vectors, which can lead to dynamic output that is not desired. To address this, we devise a first-frame-aware acyclic positional encoding (FFA-APE) method that fixes $\bm v_0$ while maintaining the smoothness of APE (see Fig.~\ref{fig:ffa-ape}). We achieve this by replacing the noise vector for the first anchor point with a learnable constant vector $\cm_0$, and using left-sided \texttt{conv1d} layers with constant padding instead of the padding-less \texttt{conv1d} layers. This way, the value of $\bm v_0$ only depends on the constant vector $\cm_0$ and the left-padded vectors, which are also constant. As a result, $\bm v_0$ is naturally fixed without affecting the continuity of positional encoding.

\subsection{FFA-ST}

In this section, we introduce the first-frame-aware sparse training specially designed for our framework.
Recent non-autoregressive video generation approaches~\cite{digan,styleganv} use a discriminator design that only considers $k$ frames $\bm x_{t_1}, \cdots, \bm x_{t_k}$ for each video, distinguishing the realness of the input conditioned on the time difference of input frames $\delta_i=t_{i+1}-t_i$. This training scheme is called sparse training. StyleGAN-V~\cite{styleganv} has analyzed the choice of $k$ and found that $k=3$ is ideal for most datasets. The discriminator is defined as $D(\bm x_{t_{1,2,3}},\delta_{1,2})$.

We follow this training scheme to make full use of our non-autoregressive framework. Nonetheless, using only three randomly sampled timestamps to train the generator and discriminator can result in sharp transitions at the beginning of the generated video, where the generated $\bm x_0$ and $\bm x_1$ usually diverge too much, and sometimes even switch to another identity and never return. This happens because although we define the generation process of a video as a modulated inversion process of the start frame, the discriminator is unaware of it. The discriminator only focuses on the smoothness of generated latent trajectories, failing to ensure the motion generator produces frames that share the identity with the start frame.

To solve this problem, we introduce the initial frame into the discriminator to enhance content consistency and motion smoothness. The adversarial loss for the first-frame-aware discriminator (FFA-D) can be written as:
\begin{equation}
    \label{eq:advloss}
    \begin{aligned}
    \bm y_{t_{0,1,2,3}} &= G(\mathrm{StyleInV}(w_0, t_{0,1,2,3})), \\
    \mathcal{L}_{adv} &= 
    {\mathbb{E}_{{\bm x\sim p_v}}}\left [\log D(\bm x_{t_{0,1,2,3}}, \delta_{0,1,2})  \right ] \\ 
    &+{\mathbb{E}_{{w_0\sim p_{\mathcal{W}}}}}\left[ \log (1 - D(\bm y_{t_{0,1,2,3}},\delta_{0,1,2}) \right ],
    \end{aligned}
\end{equation}
where we specify $t_0=0$. Here, $p_v$ and $p_{\mathcal{W}}$ denote the real data distribution and $\mathcal{W}$ latent space distribution, respectively. 
To explicitly enforce initial frame reconstruction, we use a $L_2$ loss for the generated $\bm y_{t_0}$:
\begin{equation}
    \label{eq:l2loss}
    \mathcal{L}_{L_2}=|| G(w_0) - G(\mathrm{StyleInV}(w_0, 0)) ||_2.
\end{equation}

Finally, we apply latent regularization \cite{pSp, styleclip} to the encoder's output, so as to enhance content consistency:
\begin{equation}
    \label{eq:gregloss}
    \mathcal{L}_{reg} = \sum\nolimits_{i=0}^{3} || E(G(w_0), t_i) ||_2.
\end{equation}
The overall loss function for training our motion generator and the discriminator is defined as:
\begin{equation}
    \label{eq:final}
    \min\limits_{E}\max\limits_{D} \mathcal{L}_{adv}
    + \min\limits_{E}(\lambda_{L_2}\mathcal{L}_{L_2} + \lambda_{reg}\mathcal{L}_{reg}).
\end{equation}
Here $\lambda_{L_2}$ and $\lambda_{reg}$ are the loss hyperparameters. We also apply discriminator adaptive augmentation \cite{karras2020training,styleganv} and $r1$ regularization \cite{karras2019style, styleganv} to further improve the training stability and generation quality.

\subsection{Finetuning-based Style Transfer}
\label{sec:finetune_transfer}
\begin{figure}
    \centering
    \includegraphics[width=1.0\linewidth]{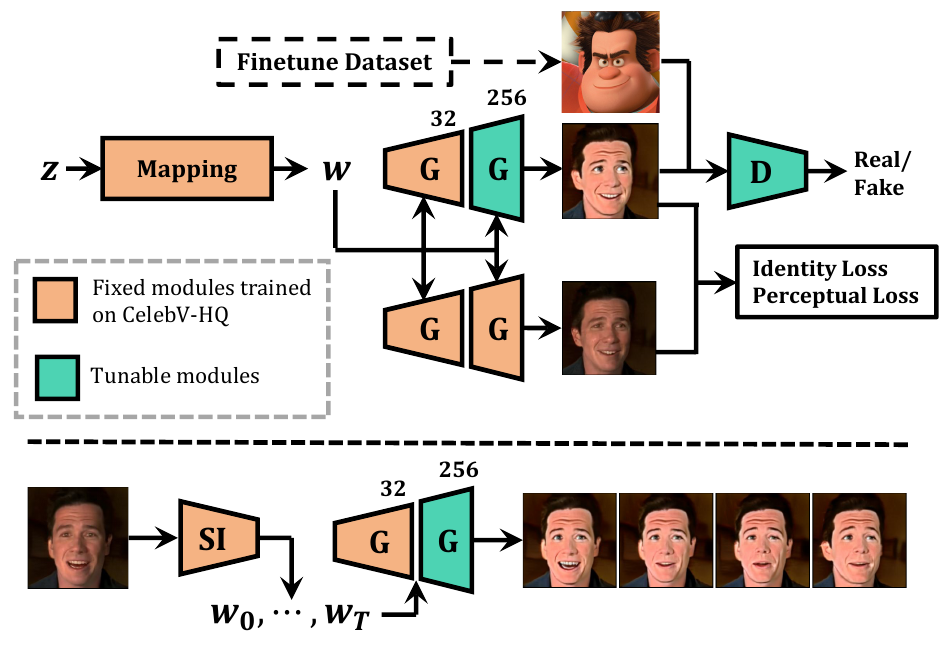}
    \caption{Finetuning-based style transfer: our framework allows easy fine-tuning of decoder (a pretrained StyleGAN generator) to a new domain by freezing mapping and low-resolution ($\leqslant32^2$) layers. Standard identity loss and perceptual loss are applied to improve identity preservation and reduce artifacts. The style-transferred videos can then be generated by incorporating the StyleInV motion generator.}
    \label{fig:transfer_pipeline}
\end{figure}

Our `inversion encoder$+$decoder' framework can naturally take a pretrained StyleGAN model as the generator. And such a configuration allows the generator to be fine-tuned for different styles, and yet still able to use the motion generator for generating new video with styles.
The capability is not possible with existing non-autoregressive video generation methods~\cite{digan, styleganv} because they cannot be finetuned under an image GAN training scheme.

To achieve style transfer, as illustrated in Fig.~\ref{fig:transfer_pipeline}, we fine-tune the pre-trained StyleGAN model using an image dataset, such as MegaCartoon~\cite{pinkney2020megacartoon}, while keeping the mapping network and low-resolution ($\leqslant32^2$, coarse and middle layers in \cite{karras2019style}) synthesis blocks fixed. 
This configuration maintains the distribution of the $\mathcal{W}$ space during fine-tuning. To improve identity preservation and reduce artifacts, we apply both a perceptual loss \cite{johnson2016perceptualloss} and an identity loss \cite{deng2019arcface} between the images generated by the original and fine-tuned StyleGAN. We show some visual results in Fig.~\ref{fig:new_style_transfer}. The style-transferred video maintains the same motion pattern as the video generated by the parent model, while adopting a new style from the fine-tuning image dataset.
It is noteworthy that the finetuning process is independent of the video generation training. It means that the finetuning-based style transfer is ``plug-and-play" as the fine-tuned image generator can be used on any StyleInV models. It does not introduce inference latency either.

\section{Experiments}
\label{sec:experiments}

\noindent\textbf{Datasets.} 
We use four video datasets in our main experiments: DeeperForensics $256^2$~\cite{jiang2020deeperforensics}, FaceForensics $256^2$~\cite{faceforensics}, SkyTimelapse $256^2$~\cite{skytimelapse} and TaiChi $256^2$~\cite{taichi}. The cropping strategy for DeeperForensics~\cite{jiang2020deeperforensics} and FaceForensics~\cite{faceforensics} is different. For DeeperForensics, we use a stabilized FFHQ~\cite{karras2019style} cropping strategy~\cite{naruniec2020high_disneyfaceswap}, while we follow the strategy firstly adopted by TGAN-V2~\cite{tganv2} for FaceForensics. Please refer to Appx.~\ref{supp:cropping} and Appx.~\ref{supp:dataset} for a detailed discussion.

\noindent\textbf{Baselines.} 
We explore four state-of-the-art methods for comparison: MoCoGAN-HD~\cite{mocoganhd}, DIGAN~\cite{digan}, StyleGAN-V~\cite{styleganv} and Long-Video-GAN~\cite{brooks2022longvideogan}. 
Among these methods, MoCoGAN-HD and DIGAN require an explicit setting of the training clip length. We follow the default setting of their paper to set the clip length as $16$ for both methods. This setting is identical to StyleGAN-V \cite{styleganv}. 

In addition, on DeeperForensics, we explore an optimized setting on DIGAN and MoCoGAN-HD for a more fair comparison. For DIGAN, we directly increase the clip length to 128 frames. For MoCoGAN-HD, we apply the first-frame-aware sparse training to train its motion generator, so as to avoid using a heavy 3D discriminator, allowing it to be trained with 128-frame clips as other methods.

\noindent\textbf{Evaluation.} 
We use Fr\'echet Inception Distance (FID) \cite{FID} and Fr\'echet Video Distance (FVD) \cite{FVD} to evaluate all models quantitatively. In practice, we follow the metric calculation framework provided by StyleGAN-V~\cite{styleganv} to first generate a fake video dataset with 2,048 synthesized clips, each of 128 frames. For FID, we sample 50k frames from real and fake video datasets to compute the result. For FVD, we compute FVD$_{16}$ and FVD$_{128}$ with the first 16 frames and all 128 frames of each clip, respectively. We use FID results to show the single-frame image quality of each method.

To ensure a fair comparison, we re-benchmark the quantitative results of every method on every dataset. We retrain all the baselines using the official paper setting, except for MoCoGAN-HD on SkyTimelapse, where an officially released checkpoint is available.
For more implementation details, please refer to Appx.~\ref{supp:training}.

\subsection{Main Results}
\label{sec:exp_main}

\begin{table}[t]
\centering\small
\caption{FID, FVD$_{16}$ and FVD$_{128}$ results of video generation methods on (a) DeeperForensics $256^2$, (b) FaceForensics $256^2$, (c) TaiChi $256^2$, and (d) SkyTimelapse $256^2$. \textbf{Bolds} indicate best and \underline{underlines} indicate the second best.}\label{tab:main}
\vspace{-0.05in}
\begin{subtable}{\linewidth}
\centering\small
\caption{DeeperForensics $256^2$}\label{tab:main_dfuf}
\begin{tabular}{lccc}
    \toprule
    Method & FID ($\downarrow$) & FVD$_{16}$ ($\downarrow$) & FVD$_{128}$ ($\downarrow$) \\
    \midrule
    MoCoGAN-HD   & 135.30 & 101.07 & 610.30 \\
    DIGAN & 191.99 & 46.69 & 1060.27 \\
    StyleGAN-V & 59.59 & \textbf{39.33} & \underline{68.81} \\
    Long-Video-GAN & \underline{56.54} & 74.77 & 169.45 \\
    StyleInV (ours) & \textbf{54.05} & \underline{41.58} & \textbf{53.93} \\
    \bottomrule
\end{tabular}
\vspace{0.05in}
\centering\small
\caption{FaceForensics $256^2$}\label{tab:main_ffs}
\begin{tabular}{lccc}
    \toprule
    Method & FID ($\downarrow$) & FVD$_{16}$ ($\downarrow$) & FVD$_{128}$ ($\downarrow$) \\
    \midrule
    MoCoGAN-HD   & 24.45 & 112.67 & 486.69 \\
    DIGAN & 151.53 & 146.62 & 1993.20 \\
    StyleGAN-V & \textbf{8.64} & \underline{52.92} & \underline{108.86} \\
    Long-Video-GAN & 40.40 & 233.26 & 567.78 \\
    StyleInV (ours) & \underline{12.06} & \textbf{47.88} & \textbf{103.63} \\
    \bottomrule
\end{tabular}
\vspace{0.05in}
\centering\small
\caption{TaiChi $256^2$}\label{tab:main_taichi}
\begin{tabular}{lccc}
    \toprule
    Method & FID ($\downarrow$) & FVD$_{16}$ ($\downarrow$) & FVD$_{128}$ ($\downarrow$) \\
    \midrule
    MoCoGAN-HD   & 73.61 & 315.03 & 622.95 \\
    DIGAN & 67.24 & 196.77 & 954.93 \\
    StyleGAN-V & \textbf{35.68} & 254.74 & \underline{477.78} \\
    Long-Video-GAN & 43.90 & \underline{248.55} & 502.65 \\
    StyleInV (ours) & \underline{41.55} & \textbf{185.72} & \textbf{328.90} \\
    \bottomrule
\end{tabular}
\vspace{0.05in}
\centering\small
\caption{SkyTimelapse $256^2$}\label{tab:main_sky}
\begin{tabular}{lccc}
    \toprule
    Method & FID ($\downarrow$) & FVD$_{16}$ ($\downarrow$) & FVD$_{128}$ ($\downarrow$) \\
    \midrule
    MoCoGAN-HD   & 251.81 & 696.58 & 4116.03 \\
    DIGAN & 32.83 & 148.08 & 269.43 \\
    StyleGAN-V & \underline{16.95} & \underline{81.32} & 197.83 \\
    Long-Video-GAN & 25.41 & 116.50 & \textbf{152.70} \\
    StyleInV (ours) & \textbf{14.32} & \textbf{77.04} & \underline{194.25} \\
    \bottomrule
\end{tabular}
\end{subtable}
\end{table}
\begin{table}[t]
\caption{FID, FVD$_{16}$ and FVD$_{128}$ results of extended experiments on DeeperForensics $256^2$. We apply sparse training to MoCoGAN-HD \cite{mocoganhd} ($\#1$) and change the preset clip length of DIGAN \cite{digan} to 128 ($\#2$). \textbf{Bolds} indicate best. (-) indicates a smaller (better) quantitative result, while (+) indicates a larger (worse) one, compared with Table \ref{tab:main_dfuf}.}
\vspace{-0.1cm}
\label{tab:quant_extended}
\resizebox{1.0\linewidth}{!}{
\centering\small
\begin{tabular}{clccc}
    \toprule
    \rowNumber{\#} & Method & FID ($\downarrow$) & FVD$_{16}$ ($\downarrow$) & FVD$_{128}$ ($\downarrow$) \\
    \midrule
    \rowNumber{1} & \cite{mocoganhd} + Sparse Training & 55.84 (-) & 54.58 (-) & 129.13 (-) \\
    \rowNumber{2} & \cite{digan} + Clip 128 & 74.80 (-) & 87.42 (+) & 95.80 (-) \\
    \rowNumber{3} & StyleInV (ours) & \textbf{54.05} & \textbf{41.58} & \textbf{53.93} \\
    \bottomrule
\end{tabular}
}
\end{table}
\begin{figure*}
    \centering
    \begin{subfigure}[b]{\linewidth}
        \centering

        \begin{subfigure}[b]{0.48\linewidth}
            \centering
            \includegraphics[width=\textwidth]{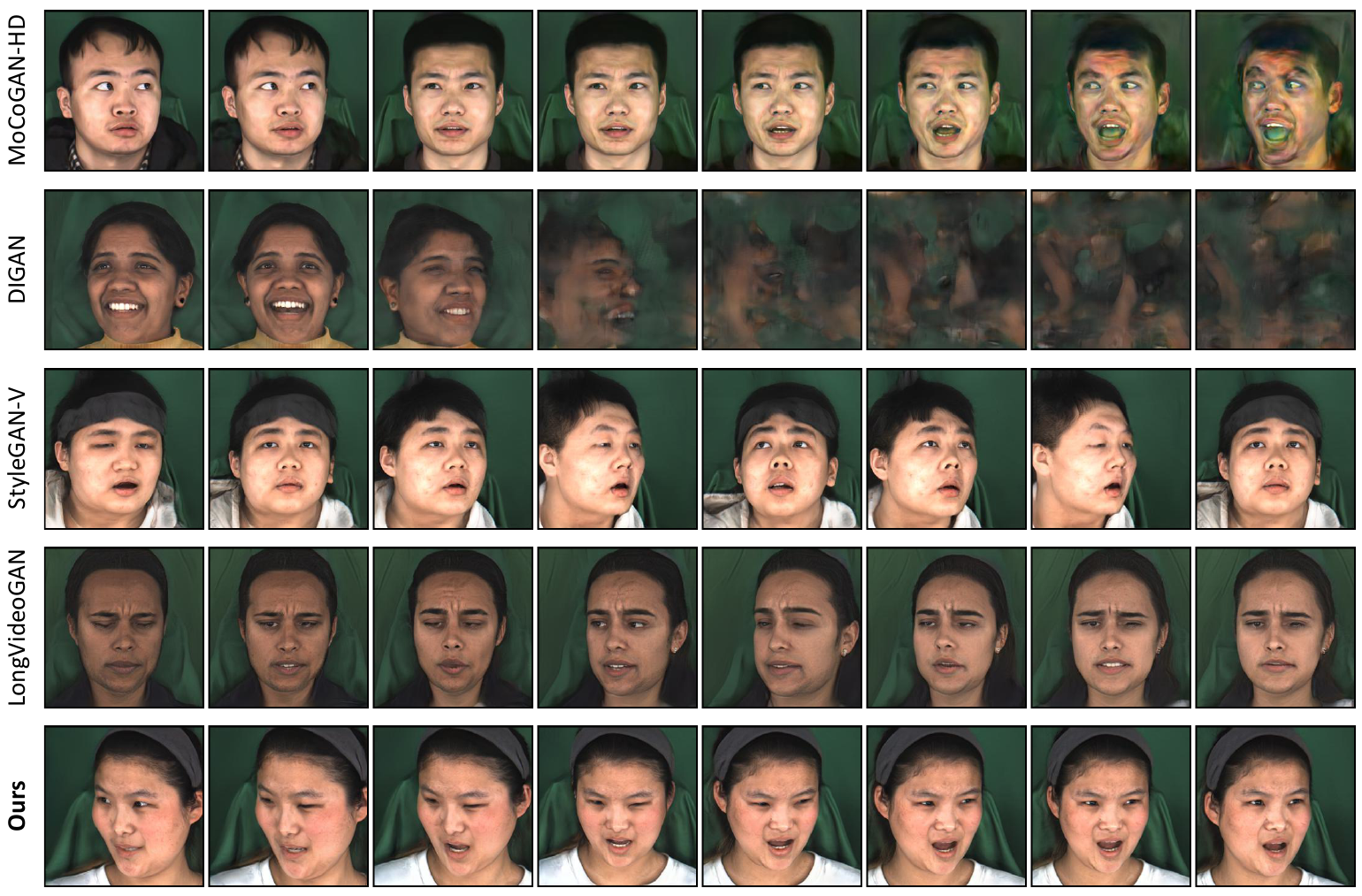}    
            \caption{DeeperForensics $256^2$}
        \end{subfigure}
        \quad
        \begin{subfigure}[b]{0.48\linewidth}
            \centering
            \includegraphics[width=\textwidth]{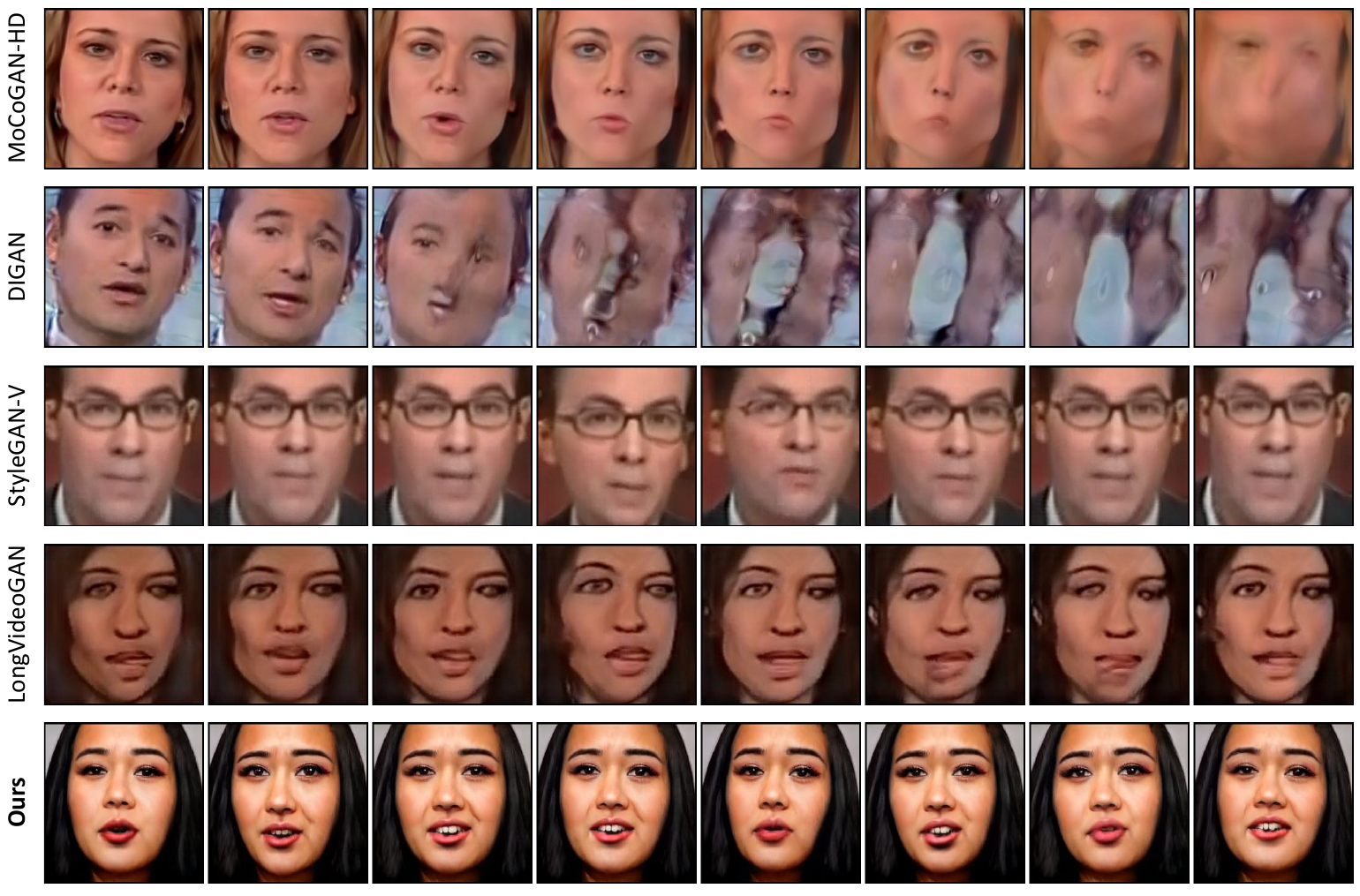}
            \caption{FaceForensics $256^2$}
        \end{subfigure}
    \end{subfigure}

    \begin{subfigure}[b]{\linewidth}
        \centering

        \begin{subfigure}[b]{0.48\linewidth}
            \centering
            \includegraphics[width=\textwidth]{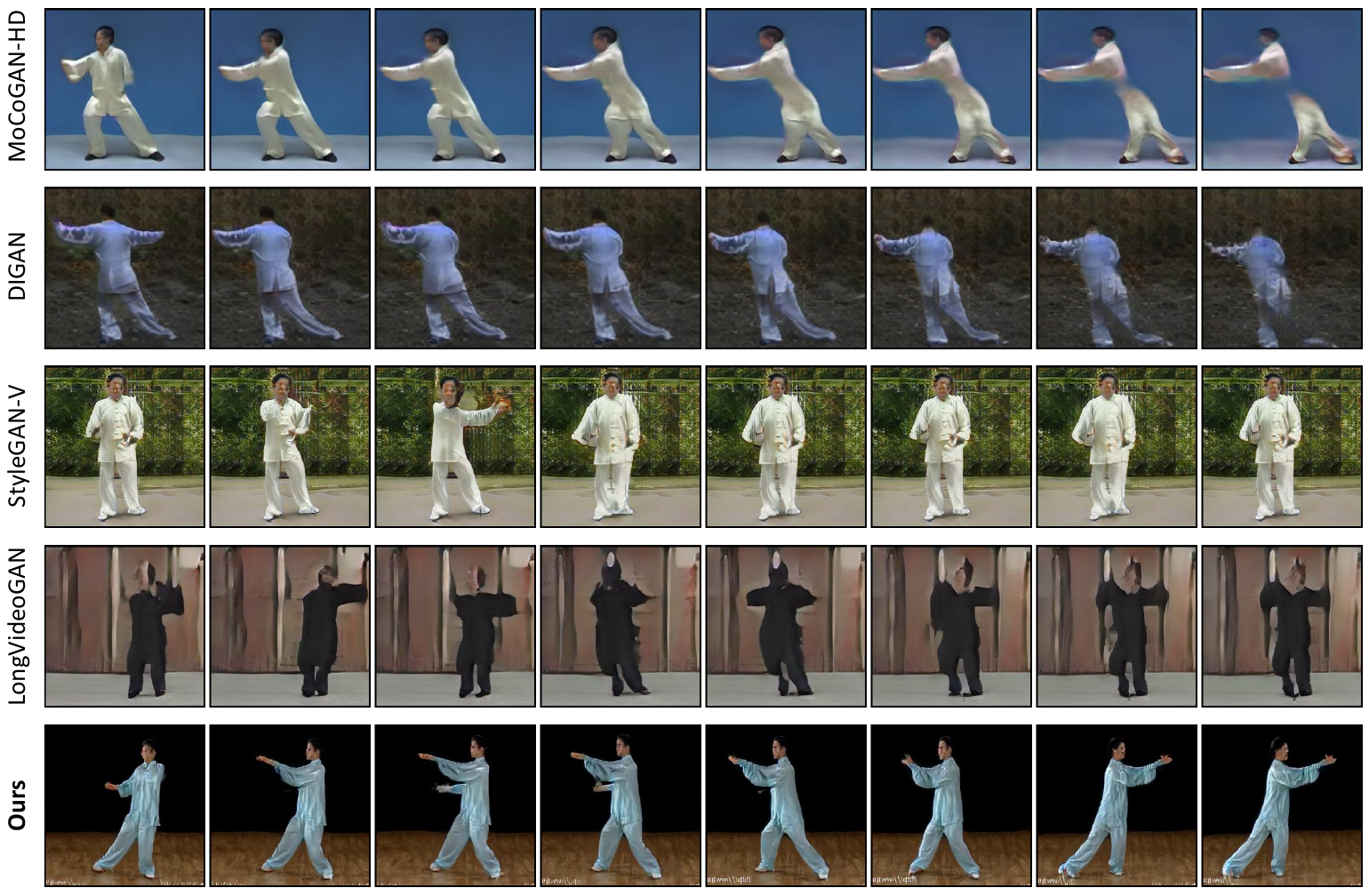}    
            \caption{TaiChi $256^2$}
        \end{subfigure}
        \quad
        \begin{subfigure}[b]{0.48\linewidth}
            \centering
            \includegraphics[width=\textwidth]{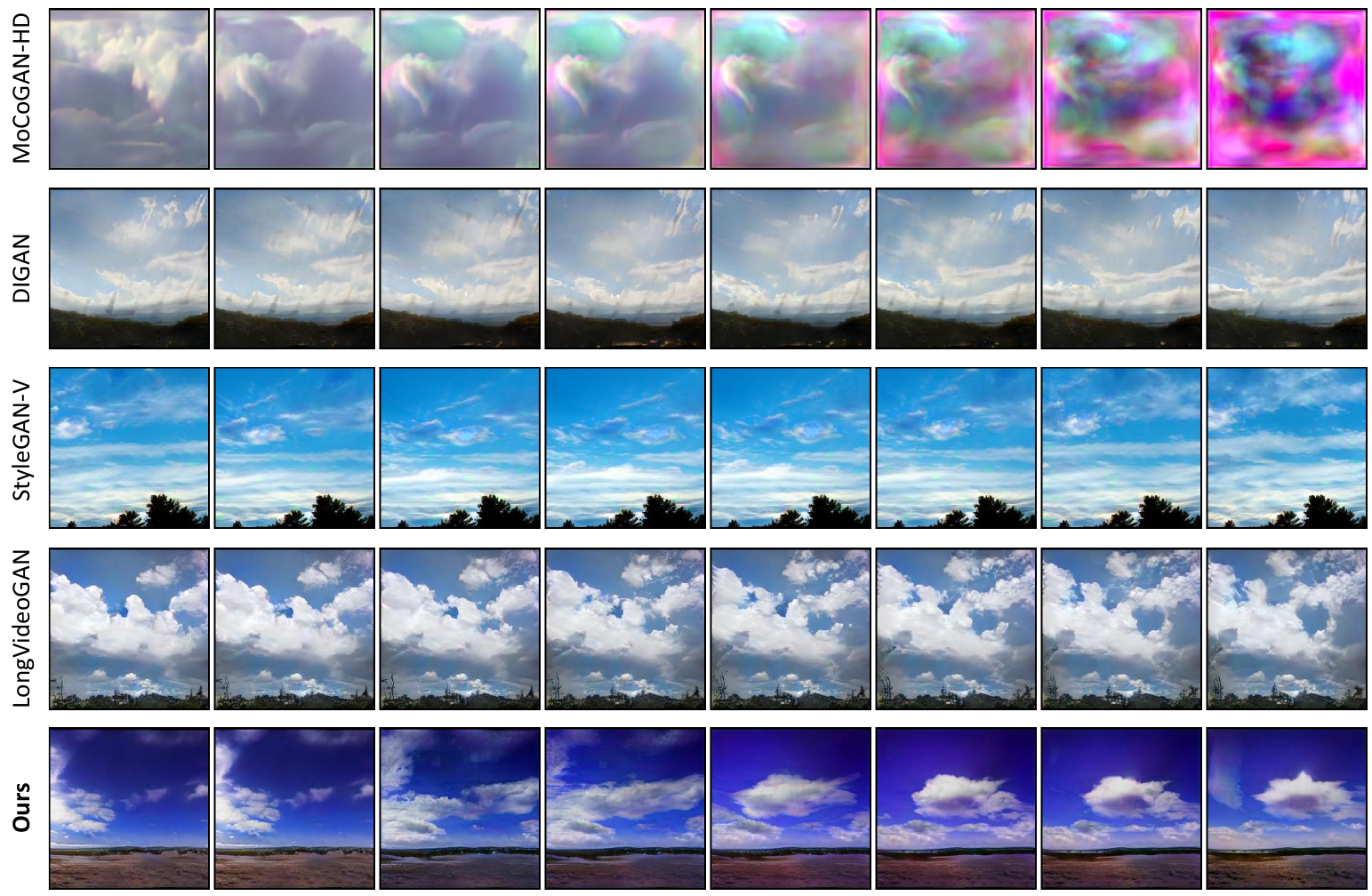}
            \caption{SkyTimelapse $256^2$}
        \end{subfigure}
    \end{subfigure}
    
    \caption{\textit{Uncurated} samples from the existing methods on DeeperForensics $256^2$, FaceForensics $256^2$, TaiChi $256^2$ and SkyTimelapse $256^2$, respectively. We sample a 128-frame video and display every 16 frames, starting from $t = 0$.}
    \label{fig:qualitative}
\end{figure*}

\begin{figure*}
    \centering
    \includegraphics[width=\linewidth]{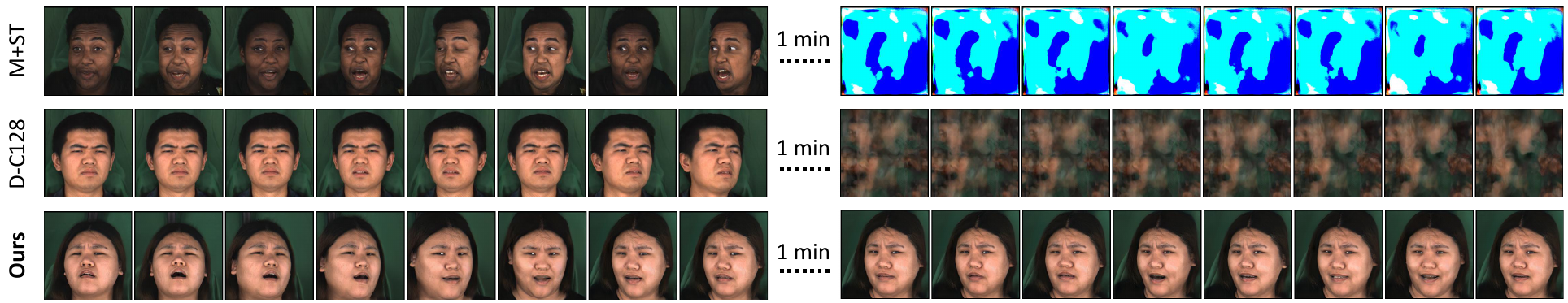}
    \caption{Qualitative comparison of extended experiments. ``M+ST" and ``D-C128" correspond to Table \ref{tab:quant_extended} ($\#1$) and ($\#2$), respectively. Each row shows the first and last 128 frames of a 2056-frame (68.5s) video, displayed every 16 frames.}
    \label{fig:qualitative_extend}
\end{figure*}

\noindent\textbf{Quantitative results.} 
Table~\ref{tab:main} summarizes the quantitative results of our method compared to other baselines. Our method achieves competitive quantitative results on all the benchmarks. Notably, although MoCoGAN-HD and DIGAN are trained with clips of 16 frames, we still outperform them in terms of FVD$_{16}$ metrics on all four datasets. 

\noindent\textbf{Qualitative results.}
Figure~\ref{fig:qualitative} shows the qualitative comparison between our method and the baselines on all four datasets. MoCoGAN-HD and DIGAN both suffer from motion collapse, resulting in a degraded generation quality over time. StyleGAN-V shows an impressive visual performance on FaceForensics and SkyTimelapse, but it sometimes fails to maintain the identity and accessories on DeeperForensics and lacks diversity and magnitude of motion over a long time span on TaiChi (the subject gradually fixes at one state). Long-Video-GAN is exceptionally good at SkyTimelapse, but it cannot achieve similar performance on other datasets. It fails to maintain the identity on DeeperForensics, and its single-frame content on TaiChi lacks details and is inferior to other methods.
The generated videos by Long-Video-GAN collapse on FaceForensics.

In contrast to existing methods, our method demonstrates stable results on all four datasets, particularly with superior identity preservation on human-face video and long-term generation quality on TaiChi. Although our method outperforms existing methods in terms of content quality, continuity, and quantitative results, the motion semantics of our generated videos on SkyTimelapse are inferior to those on other datasets. This could be one of the limitations of our work and an area for future improvement.

\begin{figure*}
    \centering
    \includegraphics[width=\linewidth]{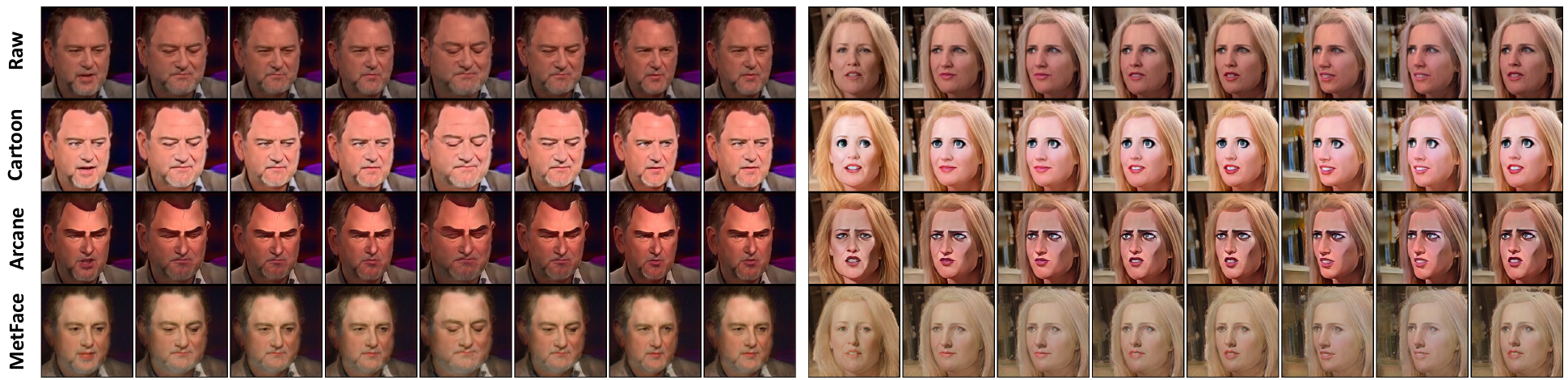}
    \caption{Finetuning-based style transfer result. The 1st row is generated by the parent model trained on CelebV-HQ \cite{zhu2022celebvhq}. The 2nd, 3rd, and 4th row uses the StyleGAN generator fintuned on Cartoon, Arcane, and MetFace, respectively.}
    \label{fig:new_style_transfer}
\end{figure*}

\noindent\textbf{Extended experiments.}
We present more in-depth comparisons in Table \ref{tab:quant_extended} and Fig.~\ref{fig:qualitative_extend} by introducing training improvements to baselines. Increasing the clip length generally improves the results of MoCoGAN-HD and DIGAN, but they are still inferior to our method. Notably, training with longer clips harms the short-term FVD$_{16}$ result of DIGAN, which indicates its tradeoff between duration length and local temporal quality. Qualitatively, for both methods, the generated content is evidently improved within 128 frames, although the sparsely trained MoCoGAN-HD exhibits issues with identity switching. Motion collapse is still observed when MoCoGAN-HD and DIGAN generates long videos. In contrast, our method can stably generate extremely long videos without motion collapse. Our method outperforms the sparsely trained MoCoGAN-HD, demonstrating the superiority of our motion generator design.

\subsection{Properties}

As discussed in Section~\ref{sec:finetune_transfer}, our method has the unique advantage over state-of-the-art methods, such as StyleGAN-V and Long-Video-GAN, on its high compatibility with StyleGAN-based downstream techniques. 

\noindent\textbf{Finetuning-based style transfer.} 
We train the parent model (motion generator and StyleGAN) on CelebV-HQ \cite{zhu2022celebvhq}, as its rich identity makes it more suitable for transfer learning. To perform style transfer, we fine-tune the StyleGAN on the Cartoon~\cite{pinkney2020megacartoon}, MetFace~\cite{karras2020training}, and Arcane datasets following the procedure outlined in Section \ref{sec:finetune_transfer}. In Fig.~\ref{fig:new_style_transfer}, we show examples where the same StyleInV-generated latent sequence is decoded by different but aligned StyleGAN generators. Our method achieves satisfactory results in terms of smooth video style transfer with well-aligned face structure, identity, and expression, demonstrating its desirable properties and potential for various applications. More results can be found on our project page.

\begin{figure}
    \centering
    \includegraphics[width=1.0\linewidth]{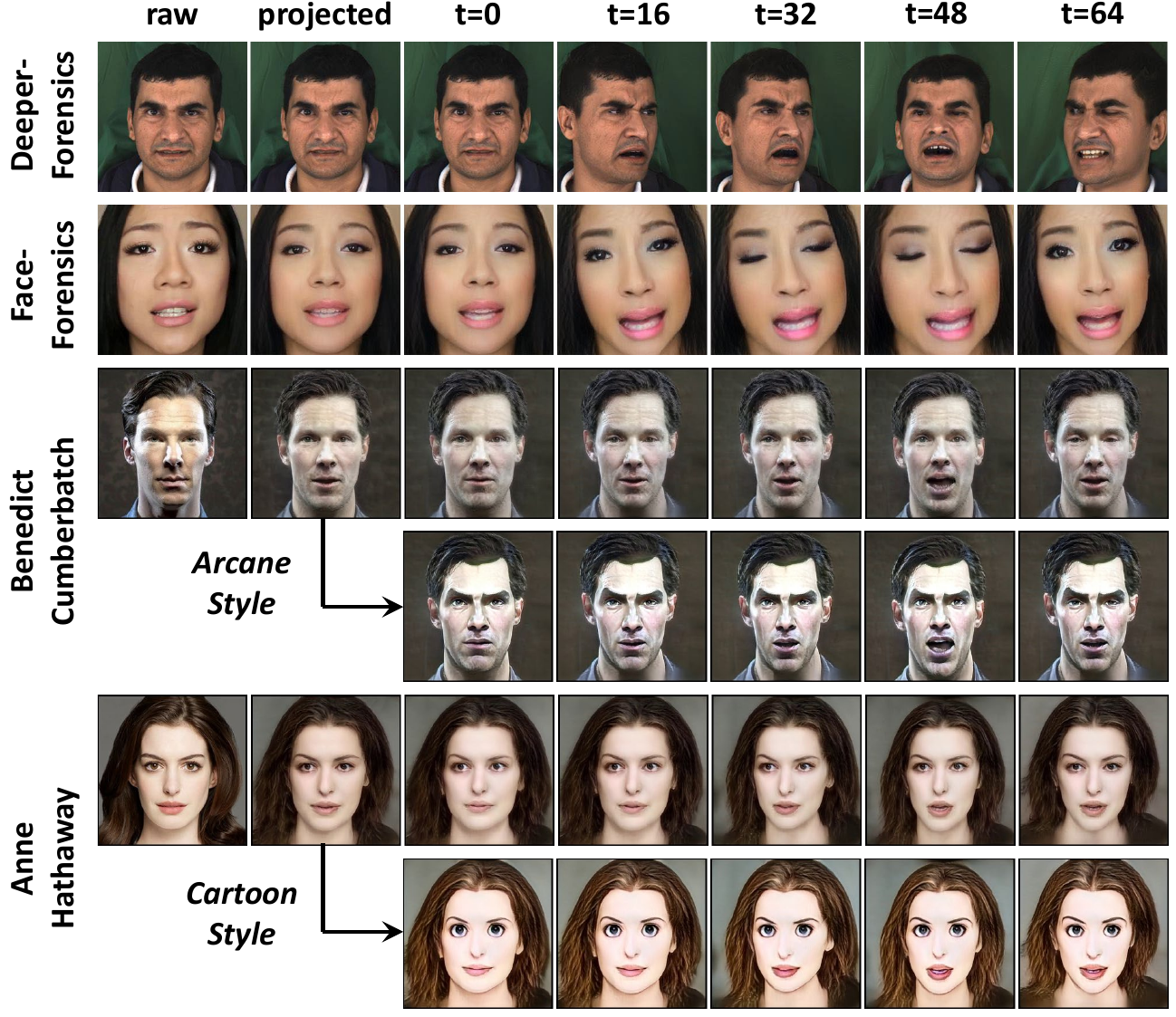}
    \caption{Initial-frame conditioned generation and style transferred results.}
    \label{fig:prop_initial}
\end{figure}
\noindent\textbf{Initial-frame conditioned generation.}
Our network supports generating a series of content given a real-world image as the initial frame. We first inverse the image into the StyleGAN2 latent space with a pSp \cite{pSp} encoder, which is trained to initialize the weights of StyleInV. We treat it as the 512-dimensional initial frame latent $w_0$, then use it to generate a video with our StyleInV. The generated latent sequence can be also applied to a finetuned image generator to synthesize a style-transferred animation video. Through this pipeline, the real image is reconstructed twice, the first time is during the inversion process, while the second time is when synthesizing $G(\mathrm{StyleInV}(w_0, 0))$.

When the real images are sampled from the training dataset (see Fig.~\ref{fig:prop_initial} first two rows), $G(\mathrm{StyleInV}(w_0, 0))$ can faithfully reconstruct the raw image and generate high-quality videos. We then test the generation quality for real images sampled out of the training set (see Fig.~\ref{fig:prop_initial} last two rows, where we select Benedict Cumberbatch and Anne Hathaway). We use the StyleGAN2 generator and StyleInV model trained on CelebV-HQ \cite{zhu2022celebvhq} dataset as it is richer in its identities. The results show that our StyleInV network can still generate meaningful videos while reconstructing the initial frame decently, and the style transfer results are smooth and well-aligned. Please refer to our project page for more results.

\subsection{Ablation Studies}

\begin{table}[t]
\caption{Ablation result on the DeeperForensics dataset.}
\label{table:ablation}
\resizebox{1.0\linewidth}{!}{
\centering \small
\begin{tabular}{llccc}
\toprule
\rowNumber{\#} & Method & FID ($\downarrow$) & FVD$_{16}$ ($\downarrow$) & FVD$_{128}$ ($\downarrow$) \\
\midrule
\rowNumber{1} & w/o inversion encoder & 54.35 & 59.49 & 152.82 \\
\midrule
\rowNumber{2} & w/o FFA-APE & 55.26 & 88.98 & 144.52 \\
\midrule
\rowNumber{3} & w/o Eq.\eqref{eq:l2loss} \& Eq.\eqref{eq:advloss} & 52.55 & 67.43 & 58.88 \\
\rowNumber{4} & w/o Eq.\eqref{eq:advloss} & 53.95 & 86.32 & 59.76 \\
\midrule
\rowNumber{5} & Ours & 54.05 & 41.58 & 53.93 \\
\bottomrule
\end{tabular}
}
\end{table}
\begin{figure}
    \centering
    \includegraphics[width=1.0\linewidth]{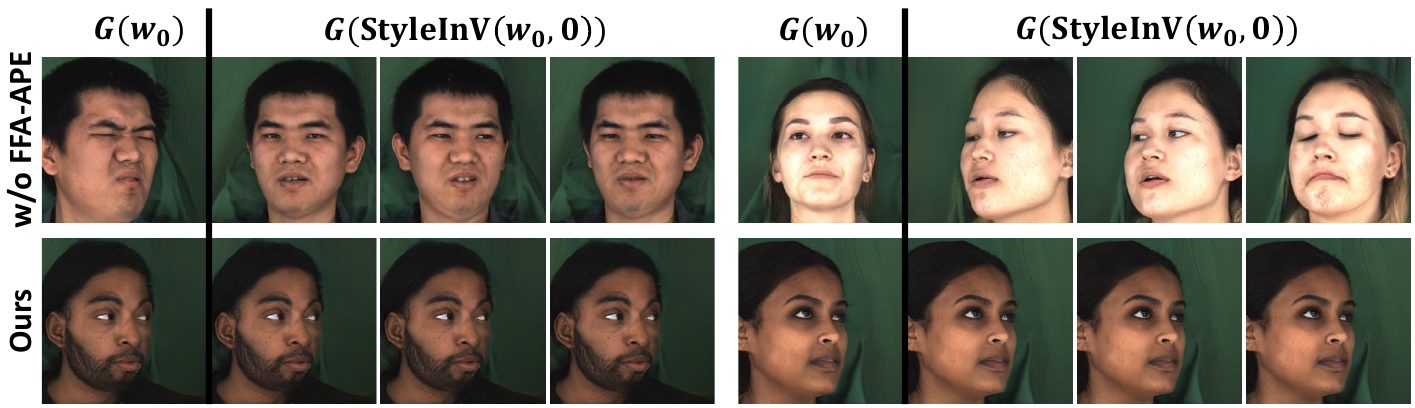}
    \caption{Ablating FFA-APE. Generate the first frame with a fixed $w_0$ but different temporal noise sequences. The model without FFA-APE fails to reconstruct the initial frame and generates the first frame with randomness.}
    \label{fig:abl_ffa_ape}
\end{figure}
\begin{figure}
    \centering
    \includegraphics[width=1.0\linewidth]{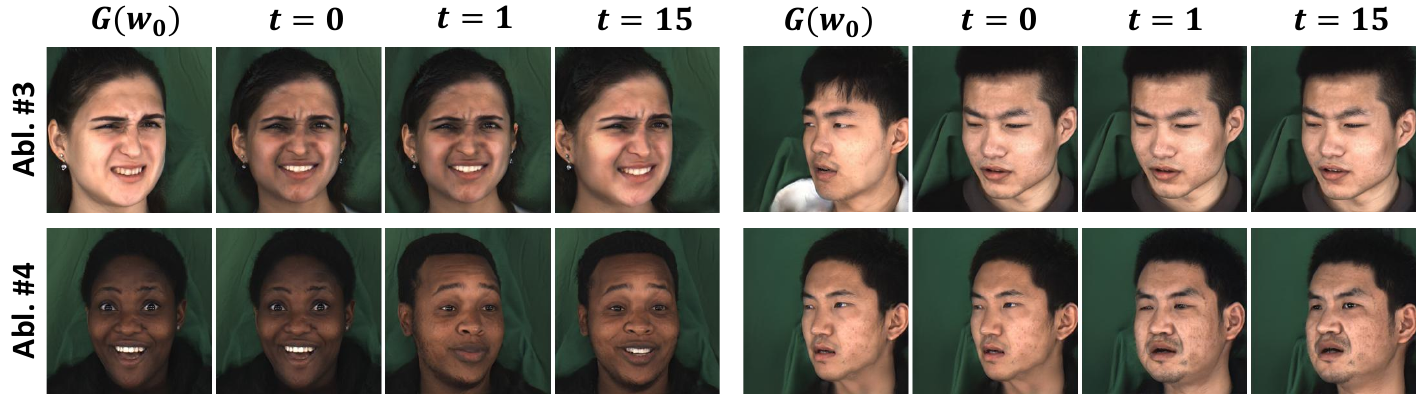}
    \caption{Ablating FFA-ST. The $L_2$ loss ensures the initial frame reconstruction. But without the initial frame included in the discriminator, we still cannot preserve the identity.}
    \label{fig:abl_ffa_st}
\end{figure}

\noindent\textbf{Motion generator design.} 
We explore two alternative motion generator designs. The first is the autoregressive MoCoGAN-HD design, which has been discussed in Section \ref{sec:exp_main}. For the second design, we remove all \texttt{Convs}, \texttt{AdaIN} and affine transform layers in Fig.~\ref{fig:pipeline} and let the mapped temporal style be the output of inversion encoder, \ie, the residual w.r.t. initial frame latent $w_0$. It largely harms identity preservation. Both FVD$_{16}$ and FVD$_{128}$ degrade significantly as is shown in Table~\ref{table:ablation}($\#1$).

\noindent\textbf{FFA-APE.}
We evaluate the importance of our first-frame-aware acyclic positional encoding (FFA-APE) module by replacing it with the original design proposed by \cite{styleganv}. As shown in Fig.~\ref{fig:abl_ffa_ape},  the dynamic embedding of the zero timestamp prevents the network from faithfully reconstructing the initial frame. In contrast, our full method can stably realize reconstruction. In addition, the L2 loss in Eq.~\eqref{eq:l2loss} fails to converge for the ablation method, and its gradient further harms the learning of the positional encoding module, leading to a much worse quantitative result shown in Table~\ref{table:ablation} ($\#2$).

\noindent\textbf{FFA-ST.} 
We conduct two ablation experiments for the FFA-ST modules. In the first experiment, we remove the initial frame from the discriminator and remove the reconstruction loss (Eq.\eqref{eq:l2loss}) (Table~\ref{table:ablation}($\#3$)). In the second experiment, we only remove the initial frame from the discriminator while keeping the reconstruction loss (Table~\ref{table:ablation}($\#4$)). As shown in Fig.~\ref{fig:abl_ffa_st}, without the reconstruction loss, our model cannot reconstruct the initial frame accurately. With the $L_2$ loss, the initial frame is reconstructed, but there is a sudden transition between the first two frames, and sometimes, the identity also changes, leading to a worse FVD$_{16}$ result. These experiments demonstrate the importance of our first-frame-aware discriminator (FFA-D).
\section{Conclusion}
\label{sec:conclusion}

We have presented a novel approach for unconditional video generation by employing a pretrained StyleGAN image generator. The proposed StyleInV motion generator generates latents in the StyleGAN2 latent space by modulating a learning-based inversion network, and thus capable of inheriting its informative priors of the initial latent. 
Our network features non-autoregressive training and uniquely supports fine-tuning based style transfer. Extensive experiments demonstrate the superiority of our method in generating long and high-resolution videos, outperforming state-of-the-art baselines. Here we also briefly discuss our limitations and broader impacts.

\subsection{Limitations}

\noindent \textbf{Inferior motion semantics on SkyTimelapse.} Our motion semantics on SkyTimelapse~\cite{skytimelapse} are inferior to those on other datasets. This could be due to different dataset characteristics, as videos in SkyTimelapse are not subject-centric and typically driven by global motions, which does not align perfectly with our model nature.

\noindent \textbf{The impact of dataset identity richness.} When the scale of facial identities in the video dataset is too small, the effects of inversion, editing, and style transfer are constrained.

\noindent \textbf{Image generation quality.} The generation quality of StyleGAN determines the performance upper bound of our method. The images generated by the StyleGAN2 models have artifacts in the background on SkyTimelapse~\cite{skytimelapse}, and lack fine details and a sense of structure on TaiChi~\cite{taichi}.

\noindent \textbf{Model training.} Our approach is two-stage, requiring 7.5 and 9 GPU days each, which is more than the 8 GPU days of StyleGAN-V~\cite{styleganv}. Despite this, StyleInV is as efficient when finetuning the hyperparameters of the video generator, since the image generator only needs to be trained once.

\subsection{Broader Impacts}

We believe that the potential of StyleInV can be further exploited. Our method can provide a natural solution towards mega-pixel level video generation and StyleGAN-based editing, and it might in return promote the research of learning-based GAN inversion methods. 

As for the negative side, StyleInV may ease the synthesis of better-quality fake videos with threats. We believe that it can be alleviated by developing more advanced falsified media detection methods or contributing larger-scale and higher-quality forgery detection datasets.

\small \noindent\textbf{Acknowledgement.} This work is supported by the National Research Foundation, Singapore under its AI Singapore Programme (AISG Award No: AISG2-PhD-2022-01-030). It is also supported under the RIE2020 Industry Alignment Fund Industry Collaboration Projects (IAF-ICP) Funding Initiative, as well as cash and in-kind contribution from the industry partner(s). We thank Shuai Yang for his help in this work.

{\small
\bibliographystyle{ieee_fullname}
\bibliography{egbib}
}

\clearpage
\section*{Appendix}
\label{sec:appendix}
This document provides supplementary information that is not elaborated in our main paper. 
In Section~\ref{supp:prop}, we present some extra properties of our method. 
In Section~\ref{supp:limit_impact}, we give a more detailed discussion of our limitations and the broader impacts.
In Section~\ref{supp:cost}, we compare the computational cost of the baselines and our model.
In Section~\ref{supp:cropping}, we introduce the different cropping strategies we applied to the DeeperForensics and FaceForensics datasets and their impact on style transfer.
In Section~\ref{supp:noise}, we show the effect of noise injection in StyleGAN models for video generation on different datasets.
In Section~\ref{supp:training}, we list the details of our model architecture and training setting.
In Section~\ref{supp:dataset}, we give a brief introduction to each dataset we use.

\appendix
\section{Other Properties}
\label{supp:prop}

Here we provide examples of other intriguing properties that our method has.

\begin{figure}
    \centering
    \includegraphics[width=1.0\linewidth]{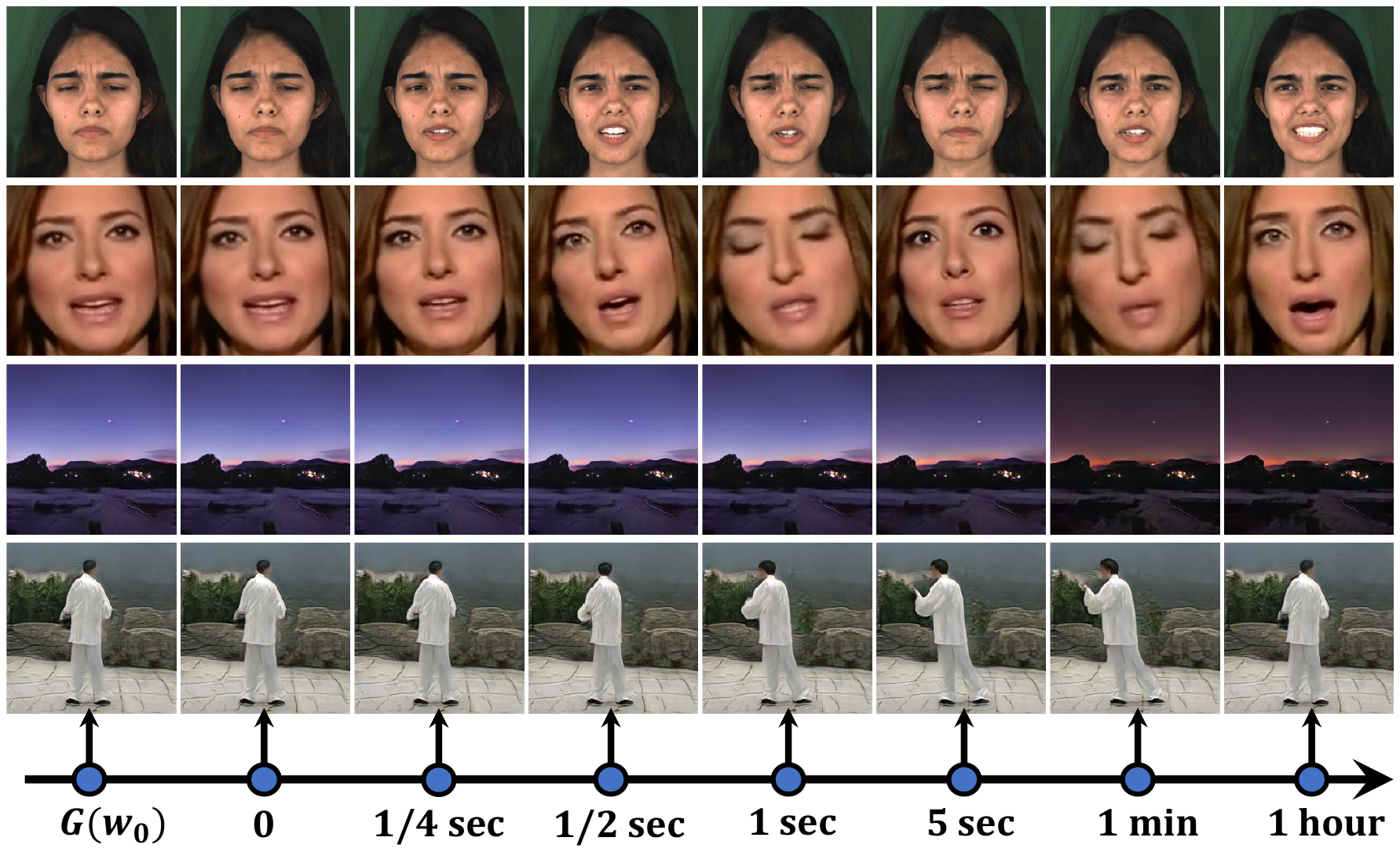}
    \caption{Our StyleInV can generate arbitrarily long videos with long-lasting content consistency.}
    \label{fig_supp:prop_ext_long}
    \vspace{-0.3cm}
\end{figure}

\vspace{0.2cm}
\noindent\textbf{Long video generation.} Similar to \cite{styleganv}, our network can also generate arbitrarily long videos with decent quality. The result is shown in Fig.~\ref{fig_supp:prop_ext_long} by extending the input timestamps to as large as one hour. Notably, our method can well preserve the content consistency of the generated videos without the motion collapse effect. Video examples are provided in the supplementary video and additional samples.

\begin{figure*}
    \centering
    \includegraphics[width=0.9\linewidth]{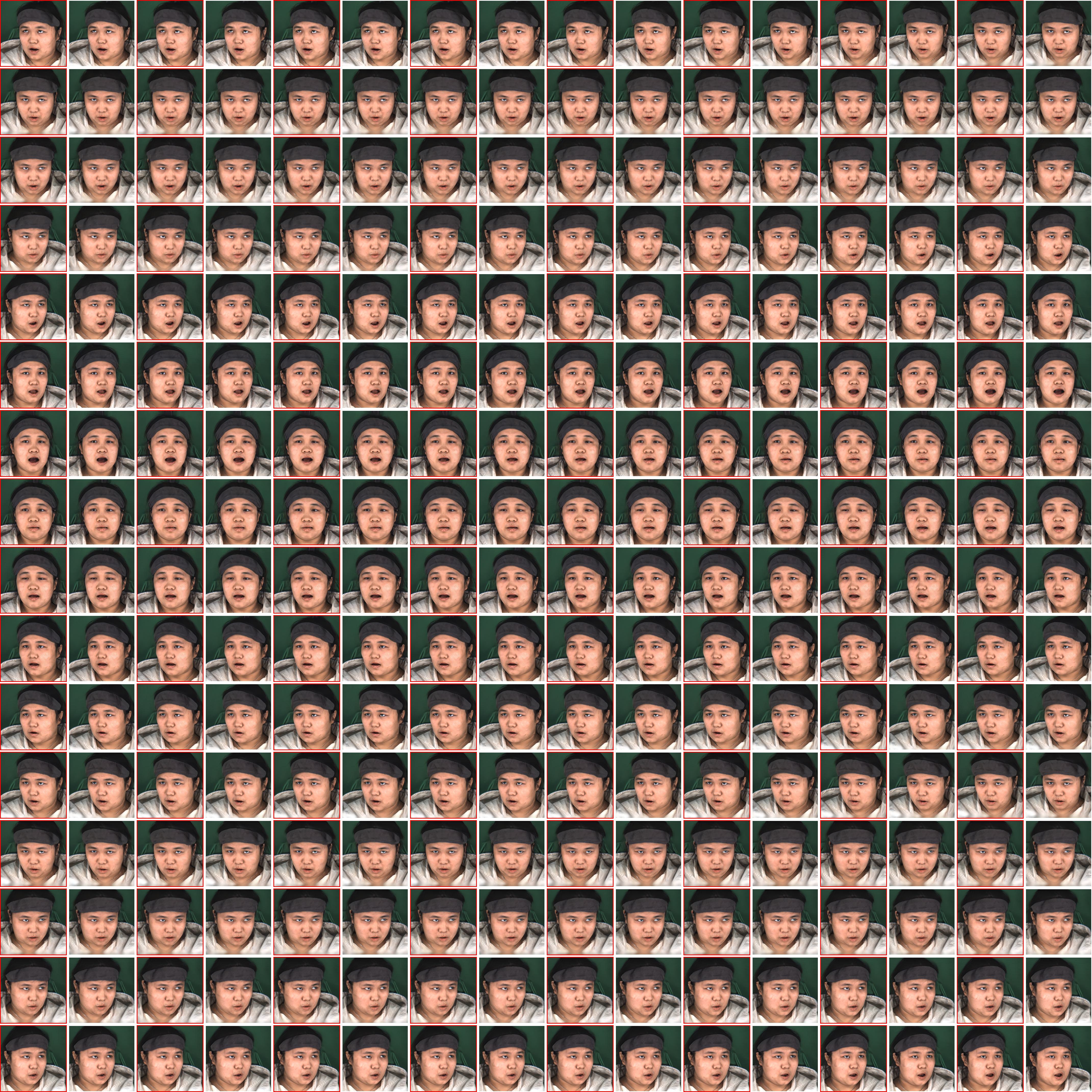}
    \caption{Temporal interpolation. All the frames with red borders form a 128-frame, 30FPS video ($\sim$4.3 seconds). The frames without borders are the interpolated ones that increase the FPS to 60 (still $\sim$4.3 seconds). View the first row first from left to right, then view the second row from left to right, then the third row, and so on.}
    \label{fig_supp:prop_interp}
\end{figure*}

\vspace{0.2cm}
\noindent\textbf{Temporal interpolation.} Our method also supports temporal interpolation to arbitrarily increase the frame rate of generated videos. Fig.~\ref{fig_supp:prop_interp} shows the result of increasing the FPS of a video from 30 to 60, by doubling the density of timestamp sampling. More specifically, for a 128-frame, 30-FPS video, we input \[t=0,1,2,\cdots,127\] to the StyleInV network, via Eq. (2) in the main paper. To increase the FPS to 60, we only need to input \[t=0,0.5,1,1.5,2,\cdots,126.5,127,127.5\], and our model can generate smooth interpolations.
\section{Limitations and Broader Impacts}
\label{supp:limit_impact}

\subsection{Limitations}

\noindent \textbf{Inferior motion semantics on SkyTimelapse.} As is mentioned in Section \ref{sec:exp_main}, the motion semantics of our generated videos on SkyTimelapse are inferior to those generated on other datasets. The reason of this may be that the characteristics of the dataset are different.  

For DeeperForensics \cite{jiang2020deeperforensics}, FaceForensics \cite{faceforensics}, and TaiChi~\cite{taichi}, the first frame largely determines the content of all frames in a video, and a video is composed of the animation process of the subject. This is consistent with the characteristics of the inversion encoder's focus on the subject. But for SkyTimelapse, two frames that are far apart often have little relation in content and the video is driven by global motions. As our network is conditioned on the first frame and predicts residuals w.r.t. the initial latent, the sky videos generated by StyleInV conform to our model nature. Please refer to the supplementary videos for visual results. 

This nature makes our model outstanding in identity preservation and can be better applied to applications like animation. Addressing more dynamics and global motions is an interesting improvement and future work for StyleInV.


\noindent \textbf{The impact of dataset identity richness.} The second limitation of our model is that, when the identity scale of the face video dataset is too small, it is difficult for us to fully inherit all the excellent properties of an FFHQ pre-trained StyleGAN2. This is why we develop our style transfer model on a recently released large-scale face video dataset CelebV-HQ \cite{zhu2022celebvhq}, as it has identity diversity on the same scale as FFHQ. Our video generation performance on CelebV-HQ demonstrates the ability of our model to generalize to larger face video datasets.

\noindent \textbf{Image generation quality.} The third limitation is that the generation quality of StyleGAN determines the performance upper bound of our method. In this work, the images generated by the StyleGAN2 models trained on SkyTimelapse and TaiChi \cite{taichi} have certain artifacts in the background. Especially for the TaiChi dataset, although our approach has greatly surpassed state-of-the-art methods in terms of quantitative metrics, the visual quality can be further improved. The generated background and human body both lack fine details and a sense of structure. That is to say, for video generation on non-face video datasets, it remains improvement space to develop a high-quality image generator.

\noindent \textbf{Model training.} Finally, our approach is two-stage and thus requires more training time compared to StyleGAN-V. Our method requires 7.5 and 9 GPU days for each stage, respectively, while StyleGAN-V is one stage and only requires 8 GPU days to train. Despite this, when finetuning hyperparameters on a dataset, our StyleInV is actually as efficient as StyleGAN-V, because the image generator only needs to be trained once and can be used for all StyleInV networks. The two stages of our method are well separated. Besides, our method has some unique properties, such as finetuning-based style transfer.

\subsection{Broader Impacts}

We believe that the potential of StyleInV can be further exploited. Our method can provide a natural solution towards mega-pixel level video generation and StyleGAN-based editing, and it might in return promote the research of learning-based GAN inversion methods. 

As for the negative side, StyleInV may ease the synthesis of better-quality fake videos that might have potential threats. We believe that this issue can be alleviated by developing more advanced falsified media detection methods or contributing larger-scale and higher-quality forgery detection datasets.

\section{Computational Cost}
\label{supp:cost}

\begin{table}[t]
\caption{GPU memory consumption of different methods for one video to be added into the batch. ``A" means autoregressive while ``N-A" means non-autoregressive. ``pSG" means employing a pretrained StyleGAN2. ``mp" stands for mixed precision. ``FpV" stands for frames per video. ``MpV" stands for memory per video, reported in GB. ``GPU Days" shows the total training time, aligned on V100 GPU.}
\vspace{0.2cm}
\label{table:computation}
\resizebox{1.0\linewidth}{!}{
\centering \small
\begin{tabular}{lcccccc}
\toprule
Method & Type & pSG & mp & FpV & MpV & GPU Days\\
\midrule
MoCoGAN-HD & A & \yes & & 16 & 5.37 & $(7.5+9)\times2$\\
MoCoGAN-HD & A & \yes & & 32 & 11.37 & $(7.5+18)\times2$\\
DIGAN & N-A & & & 2 & 1.32 & $16$ \\
StyleGAN-V & N-A & & \yes & 3 & 1.20 & $8$ \\
Long-Video-GAN & N-A &  & \yes & - & - & $16\uparrow+16\uparrow$ \\
StyleInV & N-A & \yes & \yes & 4 & 2.85 & $7.5+1+9$ \\
\bottomrule
\end{tabular}
}
\end{table}
The advantage of our method in computational cost over autoregressive approaches is mainly reflected in the GPU memory consumption during training. Table~\ref{table:computation} shows the comparison result. Our approach is the only non-autoregressive method that employs a pretrained StyleGAN generator. Our FpV is fixed and thus StyleInV can be trained on arbitrarily long videos. 

For the autoregressive MoCoGAN-HD, its memory consumption for one video in the batch is proportional to the clip length, making it difficult to be trained on long videos. Meanwhile, its codebase is $\approx2$ times slower than ours as it does not support mixed precision training. 

Compared to other non-autoregressive methods, our network consumes a bit more memory due to an extra encoder network and the initial frame included in sparse training. 

For Long-Video-GAN, its model is split into two parts, each of which requires finely setting the clip length according to the output resolution. It is also the most expensive model to train. Following its default setting, it takes 64 GPU Days to train the low-resolution model and 32 GPU days to train the high-resolution model. Due to the limitation of computing resources, we can only reduce the batch size to have each part trained in 16 GPU days, with negligible performance degradation.
\section{Cropping Strategies}
\label{supp:cropping}
In this section, we introduce the cropping strategies of the FaceForensics dataset and the DeeperForensics dataset, then explain the difference between them.

\begin{algorithm}
\caption{FaceForensics dataset cropping.}
\textbf{Input}: {$x_{min}, y_{min}, x_{max}, y_{max}$} \\
\textbf{Output}: {$\hat{x}_{min}, \hat{y}_{min}, \hat{x}_{max}, \hat{y}_{max}$}
\label{alg:faceforensics}
\begin{algorithmic}
\State $w = x_{max} - x_{min}$
\State $h = y_{max} - y_{min}$
\If{$w<h$} 
    \State $\Delta = h - w$
    \State $\hat{x}_{min} = x_{min} - \Delta/2$
    \State $\hat{x}_{max} = x_{max} + \Delta/2$
    \State $\hat{y}_{min} = y_{min}$
    \State $\hat{y}_{max} = y_{max}$
\Else
    \State $\Delta = w - h$
    \State $\hat{x}_{min} = x_{min}$
    \State $\hat{x}_{max} = x_{max}$
    \State $\hat{y}_{min} = y_{min} - \Delta/2$
    \State $\hat{y}_{max} = y_{max} + \Delta/2$
\EndIf

\end{algorithmic}
\end{algorithm}

\vspace{0.2cm}
\noindent\textbf{FaceForensics cropping.} The FaceForensics \cite{faceforensics} dataset is composed of news broadcasting videos. Apart from raw videos, it also releases labeled face masks for each frame. TGAN-V2 \cite{tganv2} proposes to crop the dataset based on these masks. For each frame, it first computes the minimum and maximum values of the coordinates of the face region to get, $x_{min}, y_{min}, x_{max}, y_{max}$. Then this rectangle region is padded to be a square, as is stated in Algorithm~\ref{alg:faceforensics}. Finally, the selected square region is cropped and resized to the target resolution to become the cropped frame.
This pipeline is followed by all recent works \cite{mocoganhd, styleganv}. We also apply it for the FaceForensics dataset pre-processing.

\begin{figure}
    \centering
    \includegraphics[width=1.0\linewidth]{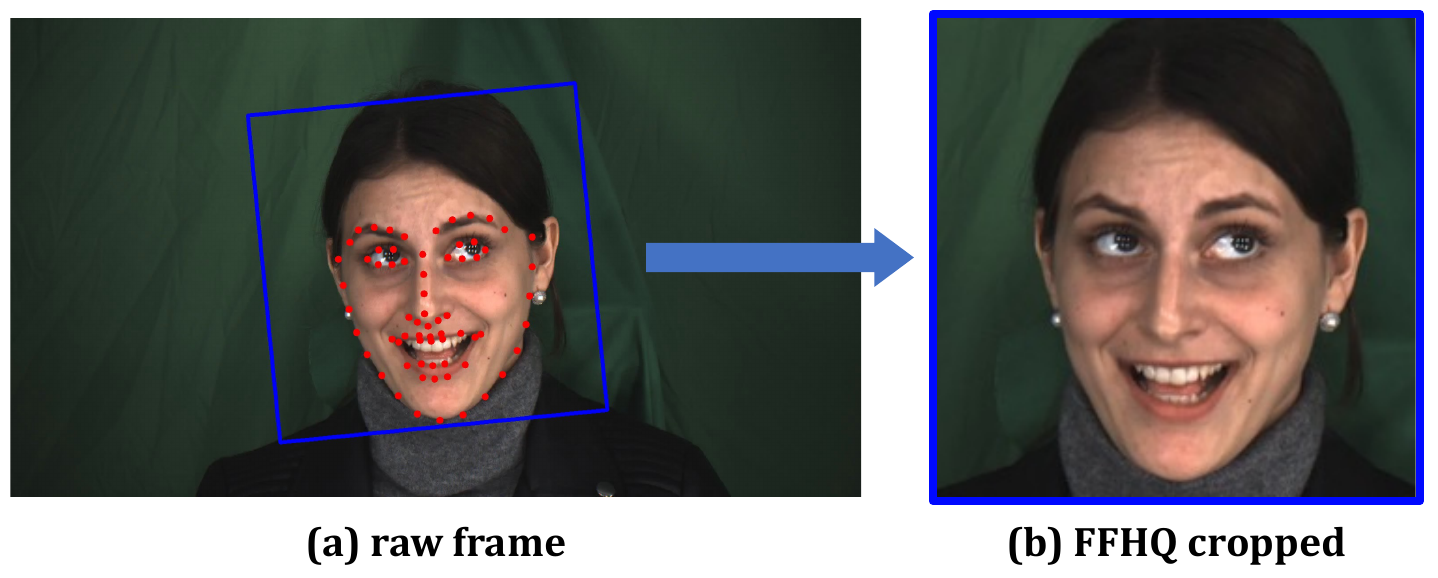}
    \caption{FFHQ cropping strategy on DeeperForensics dataset. The landmarks are detected.}
    \label{fig_supp:ffhq_crop}
\end{figure}

\vspace{0.2cm}
\noindent\textbf{DeeperForensics cropping.} The DeeperForensics \cite{jiang2020deeperforensics} dataset is composed of humans expressing given emotions. As this dataset does not release the labeled face masks, we turn to the unsupervised cropping strategy applied in FFHQ dataset \cite{karras2019style}. The cropping pipeline is shown in Fig.~\ref{fig_supp:ffhq_crop}, where the square region is determined by the detected landmarks, then the square region is resized to the target resolution.

As this cropping strategy is based on the detected landmarks, the stability of the landmark detection will greatly affect the stability of the cropped videos. In the implementation, if each frame is simply detected by a landmark detector and cropped, the cropped video will shake violently. We first replace the landmark detector with a state-of-the-art RetinaFace \cite{deng2020retinaface}, then follow a \textit{stabilizing approach} proposed by \cite{naruniec2020high_disneyfaceswap}. We find that the \textit{stabilizing approach} significantly reduces the shaking effect. Here we briefly describe it.

The state-of-the-art landmark detectors input a bounding box of the detected face and output the landmarks. We shift the bounding box at a random distance and a random angle multiple times. Then we use these bounding boxes to detect the landmarks and average the results. This approach statistically reduces the variance of the detected landmarks.

\begin{figure}
    \centering
    \includegraphics[width=1.0\linewidth]{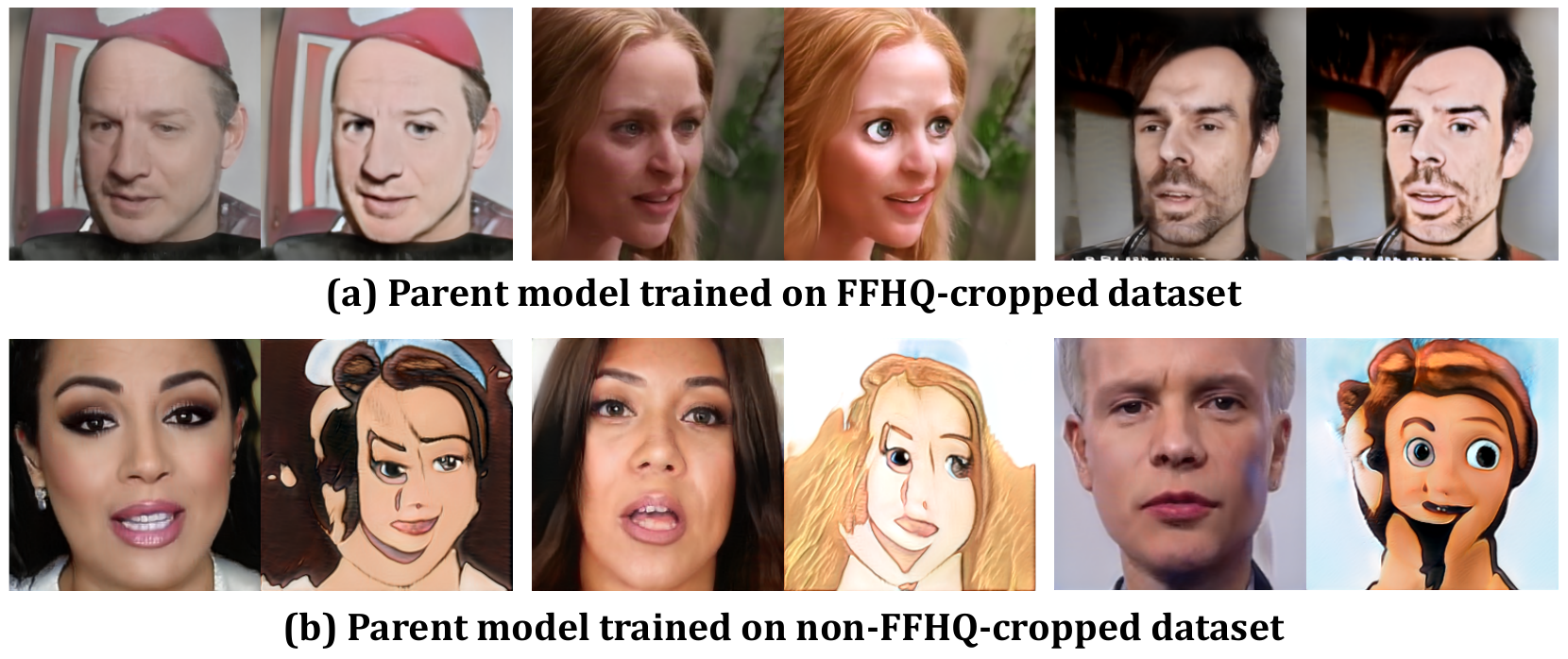}
    \caption{When the parent model is trained on an FFHQ-cropped dataset (\eg, CelebV-HQ), finetuning-based style transfer produces promising results. Otherwise, a severe structure collapse occurs.}
    \label{fig_supp:crop_finetune}
\end{figure}

\vspace{0.2cm}
\noindent\textbf{Difference.} The FFHQ cropping strategy aligns the human facial features in a fixed position. This property improves the effect of finetuning-based style transfer. As the common datasets adopted for style transfer (\eg, Cartoon \cite{pinkney2020megacartoon} and Metfaces \cite{karras2020training}) are also aligned by the FFHQ cropping strategy, when the datasets are well aligned in structure, the finetuning process can more naturally adjust the weights of high-resolution layers upon fixed low-resolution layers. Fig.~\ref{fig_supp:crop_finetune} compares the finetuning-based style transfer result of the parent model trained on CelebV-HQ \cite{zhu2022celebvhq} (where we also apply the stabilized FFHQ cropping) and FaceForensics. When the parent model is trained on a dataset (\eg, FaceForensics) which does not share the alignment of finetuning dataset (\eg, Cartoon), the style transfer fails due to the structure collapse.

\section{Effect of Noise Injection}
\label{supp:noise}

StyleGAN series \cite{karras2019style, karras2020analyzing, karras2020training} proposes to inject noise vectors at all layers of the generator for finer details in the background, hair, skin, etc. As reported by \cite{karras2019style}, the omission of noise will lead to a ``featureless painterly look". However, though designed on top of StyleGAN2, StyleGAN-V \cite{styleganv} turns off the noise injection by default for training and inference on all datasets. It also makes sense as the totally randomized noise will bring content inconsistency among frames.

In this work, we find that the effect of noise injection in our system can be different on different datasets, positive or negative. We first investigate its effect on the image generator in terms of the FID metric. On FaceForensics and TaiChi datasets, the FID results of models with or without noise are close. But on the SkyTimelapse dataset, the model without noise injection has a much better FID result.

Then we look into how the noise in the StyleGAN2 generator affects the video generation quality. The first intuitive observation is that we should apply \textbf{constant noise} for all frames when synthesizing a video, instead of injecting random noises for different frames. This is to avoid content inconsistency. Then we compare the results of StyleInV networks with or without noise. The results are exactly the opposite for the first two datasets and the third dataset. On FaceForensics and TaiChi datasets, injecting constant noise improves the FVD results significantly, while on the SkyTimelapse dataset, the model without noise gives a much better result.

We deduce that this is because there is no distinction between subject and background on the SkyTimelapse dataset, making it difficult to clarify the way the injected noise works. While on a dataset with clear subjects and backgrounds, the injected noise effectively handles the generation of stochastic aspects, leaving the latent space focusing on synthesizing the subject, which helps our StyleInV encoder find meaningful trajectories in the latent space.

\begin{table}
\caption{FID results of StyleGAN2 generator with or without noise injection.}
\label{tab_supp:noise_fid}
\resizebox{1.0\linewidth}{!}{
\centering \small
\begin{tabular}{lccc}
\toprule
Method & FaceForensics & TaiChi & SkyTimelapse \\
\midrule
with noise & 10.19 & \textbf{38.1} & 15.05 \\
w/o noise & \textbf{9.52} & 38.37 & \textbf{11.80} \\
\bottomrule
\end{tabular}
}
\end{table}
\begin{table}
\caption{FVD results of StyleInV video generator with or without noise injection in its StyleGAN2 image generator.}
\label{tab_supp:noise_fid}
\resizebox{1.0\linewidth}{!}{
\centering \small
\begin{tabular}{lcccccc}
\toprule
\multirow{2}{*}{Method} & \multicolumn{2}{c}{FaceForensics} & \multicolumn{2}{c}{TaiChi} & \multicolumn{2}{c}{SkyTimelapse} \\
& FVD$_{16}$ & FVD$_{128}$ & FVD$_{16}$ & FVD$_{128}$ & FVD$_{16}$ & FVD$_{128}$ \\
\midrule
with noise & \textbf{47.88} & \textbf{103.63} & \textbf{185.72} & \textbf{328.90} & 115.68 & 266.67 \\
w/o noise & 106.42 & 238.93 & 326.60 & 583.60 & \textbf{77.04}  & \textbf{194.25} \\
\bottomrule
\end{tabular}
}
\end{table}

\section{Implementation Details}
\label{supp:training}

In this section, we discuss the training of baselines and our model, the architecture parameters, and the detailed training setting. 

\vspace{0.2cm}
\noindent \textbf{Baseline details.} MoCoGAN-HD~\cite{mocoganhd} designs motion generators for a pretrained StyleGAN2 as we do. DIGAN~\cite{digan} and StyleGAN-V~\cite{styleganv} train the entire framework as a whole in a non-autoregressive manner. Long-Video-GAN~\cite{brooks2022longvideogan} is split into a low-resolution stage and a high-resolution stage.

All baselines are trained on 4 NVIDIA Tesla A100 GPUs. The StyleGAN2 generator for MoCoGAN-HD is pretrained with all frames of the video dataset. Then the motion generator is trained for $100$ epochs following its default setting. DIGAN models are trained under its default config for approximately four days. All StyleGAN-V models are trained under its paper setting except on DeeperForensics dataset, for which we need to increase the R1 $\gamma$ parameter by 10 times to avoid training collapse.

\vspace{0.2cm}
\noindent \textbf{Development and training.} Our StyleInV is built upon the official \texttt{PyTorch} implementation of StyleGAN2-ADA~\cite{karras2020training}, with which we enable the mixed precision setting for StyleGAN2 and significantly speed up the training. The StyleGAN2 image generator is firstly trained on all frames of the video dataset with class-aware sampling~\cite{shen2016relay, zhou2020bbn}. The noise injection is turned off for SkyTimelapse dataset only. Then we train an inversion encoder based on Fig.~3 and Eq.~(1) to initialize the convolution layers of the StyleInV encoder. Finally, the entire StyleInV model is trained under the objective of Eq.~(6). Three steps take roughly 7.5, 1, and 9 GPU days, respectively. All StyleInV models are trained on 8 NVIDIA Tesla A100 GPUs. We apply an unbalanced learning rate setting for the Adam optimizer \cite{kingma2014adam}, where the learning rate for the StyleInV encoder and the discriminator is 0.0001 and 0.002, respectively. 

\vspace{0.2cm}
\noindent\textbf{Model details.} For the computation of temporal styles, the sampled temporal noise for each timestamp is a 512-dimensional vector. FFA-APE consists of two left-sided 1D-convolution layers with kernel size 6 and padding 5. The length of the vector sequence remains unchanged after each 1D-convolution layer. The learnable interpolation part is identical to that of StyleGAN-V \cite{styleganv}. The dimension of positional encoding $v_t$ is 512. It is concatenated with the initial frame latent $w_0$ and goes through two fully connected layers to output the final temporal style, whose dimension is also 512.

For the modulated inversion encoder, its convolution blocks are identical to those in pSp inversion encoder \cite{pSp}, which compose a ResNet-50 backbone \cite{resnet}. The \texttt{AdaIN} layers are adopted from StarGAN-V2 \cite{starganv2}, with residual connection and variance normalization enabled. The \texttt{AdaIN} layers do not down-sample the feature maps. A fully connected layer is appended after the last adaptive average pooling layer to output a 512-dimensional vector, which is the residual w.r.t. $w_0$ by definition.

For the discriminator design, we simply follow the model architecture of the StyleGAN-V discriminator. We did not delve into this part. The first frame used in the discriminator is $G(w_0)$, instead of $G(\mathrm{StyleInV}(w_0, 0))$

\vspace{0.2cm}
\noindent\textbf{Training details.} For hyper-parameters of FFA-ST, we set $\lambda_{L_2}=10$ and $\lambda_{reg}=0.05$ for all four datasets. We apply adaptive differentiable augmentation \cite{karras2020training}, where the augmentation operation is always identical for all frames in a video. We use the \texttt{bgc} augmentation pipe. The augmentation target is 0.6. The R1 $\gamma$ parameter for $r1$ regularization is 1. The learning rate for the modulated inversion encoder is 0.0001. The learning rate for the discriminator is 0.002.

For the inversion encoder training which is used for weight initialization, we follow all the training settings described in the pSp paper \cite{pSp}, except that the ID loss is turned off for TaiChi and SkyTimelapse datasets.

For the finetuning-based style transfer, we fix the mapping network and synthesis layers whose resolution is no larger than 32. The training setting is identical to that of the parent model. The finetuning process takes only 4-8 GPU hours.

\section{Dataset Details}
\label{supp:dataset}

We provide dataset details in this section. 

\vspace{0.2cm}
\noindent\textbf{DeeperForensics \cite{jiang2020deeperforensics}.} This dataset is composed of 100 identities expressing eight emotions (angry, contempt, disgust, fear, happy, neutral, sad, and surprise). The videos are collected under nine lighting conditions and seven camera positions, among which we only select the condition where the lighting is uniform and the camera shoots from the straight front. All videos are cropped to 256 resolution following the stabilized FFHQ cropping strategy which is described in Section~\ref{supp:cropping}. The entire dataset has 732 videos of 194,770 frames.

\vspace{0.2cm}
\noindent\textbf{FaceForensics \cite{faceforensics}.} We follow the same cropping strategy of StyleGAN-V to process and organize the dataset. The entire dataset has 704 videos of 364,017 frames.

\vspace{0.2cm}
\noindent\textbf{SkyTimelapse \cite{skytimelapse}.} StyleGAN-V releases its SkyTimelapse $256^2$ dataset \footnote{\href{https://disk.yandex.ru/d/7JU3c5mdWQfrHw}{https://disk.yandex.ru/d/7JU3c5mdWQfrHw}}. We directly use it for our experiments. The entire dataset has 2,114 videos of 1,168,920 frames. Notably, some videos in SkyTimelapse are hours long. We use class-aware sampling in both training and metric calculation, following StyleGAN-V.

\vspace{0.2cm}
\noindent\textbf{TaiChi \cite{taichi}.} We follow the link \footnote{\href{https://github.com/AliaksandrSiarohin/first-order-model}{https://github.com/AliaksandrSiarohin/first-order-model}} provided by DIGAN to download and crop the dataset. The original dataset resolution after processing is 256, so we directly use it for all experiments. Notably, some of the video links had expired when we were processing this dataset, thus the composition of our dataset may be slightly different from previous work. The entire dataset has 3,103 videos of 951,533 frames.

\vspace{0.2cm}
\noindent\textbf{CelebV-HQ \cite{zhu2022celebvhq}.} We download the video dataset using the link for processed CelebV-HQ videos \footnote{\href{https://github.com/CelebV-HQ/CelebV-HQ/issues/8}{https://github.com/CelebV-HQ/CelebV-HQ/issues/8}} and crop the dataset to 256 resolution with stabilized FFHQ cropping. The entire dataset has 35663 videos.

\end{document}